%% file: main.tex
\documentclass[preprint,12pt]{elsarticle}

\usepackage{amssymb}
\usepackage{amsmath}
\usepackage{microtype}
\usepackage{hyperref}
\usepackage[capitalize]{cleveref}
\usepackage{url}
\usepackage[dvipsnames]{xcolor}      
\usepackage{tikz}
\usetikzlibrary{spy}
\usetikzlibrary{matrix}
\usepackage{tabularx}
\usepackage{booktabs}
\usepackage{multirow}
\usepackage{subcaption}
\usepackage{makecell}
\usepackage{adjustbox}
\usepackage{enumitem}

\usepackage{colortbl}
\colorlet{colorFst}{green!40}
\colorlet{colorBad}{red!40}
\newcommand{\fs}{\bf}

\usetikzlibrary{positioning}
\usetikzlibrary{arrows.meta, fit, backgrounds}

\definecolor{safetyColor}{HTML}{7ac74f}
\definecolor{biodiversitycolor}{HTML}{fb4d3d}
\definecolor{3dmodelColor}{HTML}{345995}
\definecolor{applicationColor}{HTML}{eac435}
\definecolor{backgroundColor}{HTML}{2a9d8f}
\definecolor{challengeColor}{HTML}{606c38}
\definecolor{applicationBackground}{HTML}{ffb703}

\journal{Advanced Engineering Informatics}

\begin{document}

\begin{frontmatter}



    \title{ThermoNeRF: Joint RGB and Thermal Novel View Synthesis for Building Facades using Multimodal Neural Radiance Fields}

    \author[1]{Mariam Hassan}
    \author[1]{Florent Forest}
    \author[1]{Olga Fink}
    \author[2]{Malcolm Mielle}

    \affiliation[1]{organization={Intelligent Maintenance and Operations Systems, EPFL},
        city={1015 Lausanne},
        country={Switzerland}}
    \affiliation[2]{organization={Schindler EPFL Lab},
        city={1015 Lausanne},
        country={Switzerland}}

    \begin{abstract}
        Thermal scene reconstruction holds great potential for various  applications, such as analyzing building energy consumption and performing non-destructive infrastructure testing.
        However, existing methods typically require dense scene measurements and often rely on RGB images for 3D geometry reconstruction, projecting  thermal information post-reconstruction. This can lead to inconsistencies  between the reconstructed geometry and  temperature data and their actual values.
        To address this challenge, we propose ThermoNeRF, a novel multimodal approach based on Neural Radiance Fields that jointly renders new RGB and thermal views of a scene, and ThermoScenes, a dataset of paired RGB+thermal images comprising 8 scenes of building facades and 8 scenes of everyday objects.
        To address the lack of texture in thermal images, ThermoNeRF uses paired RGB and thermal images to learn scene density, while separate networks estimate color and temperature data.
        Unlike comparable studies, our focus is on temperature reconstruction and experimental results demonstrate that ThermoNeRF achieves  an average mean absolute error of $1.13^\circ$C and $0.41^\circ$C for  temperature estimation in buildings and other scenes, respectively, representing an improvement of over 50\% compared to using concatenated RGB+thermal data as input to a standard NeRF.
        \href{https://github.com/Schindler-EPFL-Lab/thermo-nerf}{\url{Code}} and \href{http://dx.doi.org/10.5281/zenodo.10835108}{\url{dataset}} are available online.
    \end{abstract}

    \begin{graphicalabstract}

        \resizebox{0.95\textwidth}{!}{\input{images/summary.tikz}}

    \end{graphicalabstract}

    \begin{highlights}
        \item We propose ThermoNeRF, a method to synthesize novel RGB+thermal views.
        \item We introduce ThermoScenes, a novel dataset with 16 scenes (with 8 building facades).
        \item We show that separating RGB and thermal fields prevents information leakage.
        \item Joint learning of the density improves density estimations for both modalities.
    \end{highlights}

    \begin{keyword}

        NeRF, thermography, thermal imaging, building, novel view synthesis, building information modeling

    \end{keyword}

\end{frontmatter}


\input{introduction}

\input{related_work}
\input{method}

\input{dataset}

\input{evaluation}

\section{Conclusion}

In this work, we propose ThermoNeRF, a novel multimodal approach that leverages NeRF to render novel RGB and thermal views of a scene. Our method facilitates thermal analysis of buildings by using a sparse set of images captured from various angles  around the structure.
We curated a new dataset specifically tailored for RGB+thermal scene reconstruction, and our experimental results  show that ThermoNeRF excels at synthesizing thermal images. It achieves an average MAE of 0.41°C on indoor scenes and 1.13°C on buildings, representing respective improvements of over  55\% and 48\% compared to the baseline approach that concatenates RGB and thermal modalities. A potential limitation of our method is its reliance  on paired RGB and thermal images, as collecting such data requires precise camera calibration  to ensure accurate alignment between the two modalities, which can pose practical challenges. To address this, future work will focus on developing techniques  to train ThermoNeRF using unpaired RGB and thermal image data. We believe our method paves the way for  new opportunities in thermal building reconstruction.  For instance, future models could  analyze ThermoNeRF's outputs to detect failures in HVAC systems, piping, or electrical networks. We leave  these exciting possibilities for future exploration.

\section*{Acknowledgements}

We sincerely thank Jiarui Yu and Juliette Parchet for helping with the data collection and for reporting the results of ThermoNeRF on those datasets.
We would also like to thank Ali Waseem for running some of the experiments.

This research was supported by Innosuisse  - Swiss Innovation Agency, Innosuisse Grant n$^o$ \textbf{105.237.1 IP-ICT} titled \textbf{Insulated: integrated solution for lean and abridged thermal evaluation with digital twins} in collaboration with Schindler AG.

\bibliographystyle{elsarticle-num-names}
\bibliography{references}
\end{document}

%% file: images/summary.tikz
\tikzset{%
    dblarw/.style n args={3}{
            -latex,
            line width=#2,
            draw=black,  
            color=black, 
            opacity=0.9,
            postaction={
                    draw=#1,
                    color=#1,
                    line width=#2-#3,
                    shorten >=2*#3,
                    shorten <=#3,
                },
            dblarw/.default={gray}{5pt}{1pt},
            dblarw/.initial={gray}{5pt}{1pt},
        },
    category box/.style={
            rectangle,
            rounded corners,
            draw=gray!30,
            fill=gray!5,
            inner sep=5pt
        },
}

\begin{tikzpicture}[scale=1.15]

    \node[align=center, font=\bfseries] (title) at (0, 2.5) {ThermoNeRF};

    \begin{scope}[local bounding box=training_set]
        \node[align=center, below=0.2cm of title, font=\bfseries, text width=7.2cm] (training) {Train multimodal NeRF model\\on sparse RGB+thermal images};

        \node[inner sep=0pt] (merged) at (0,-0.5)
        {\includegraphics[width=.2\textwidth]{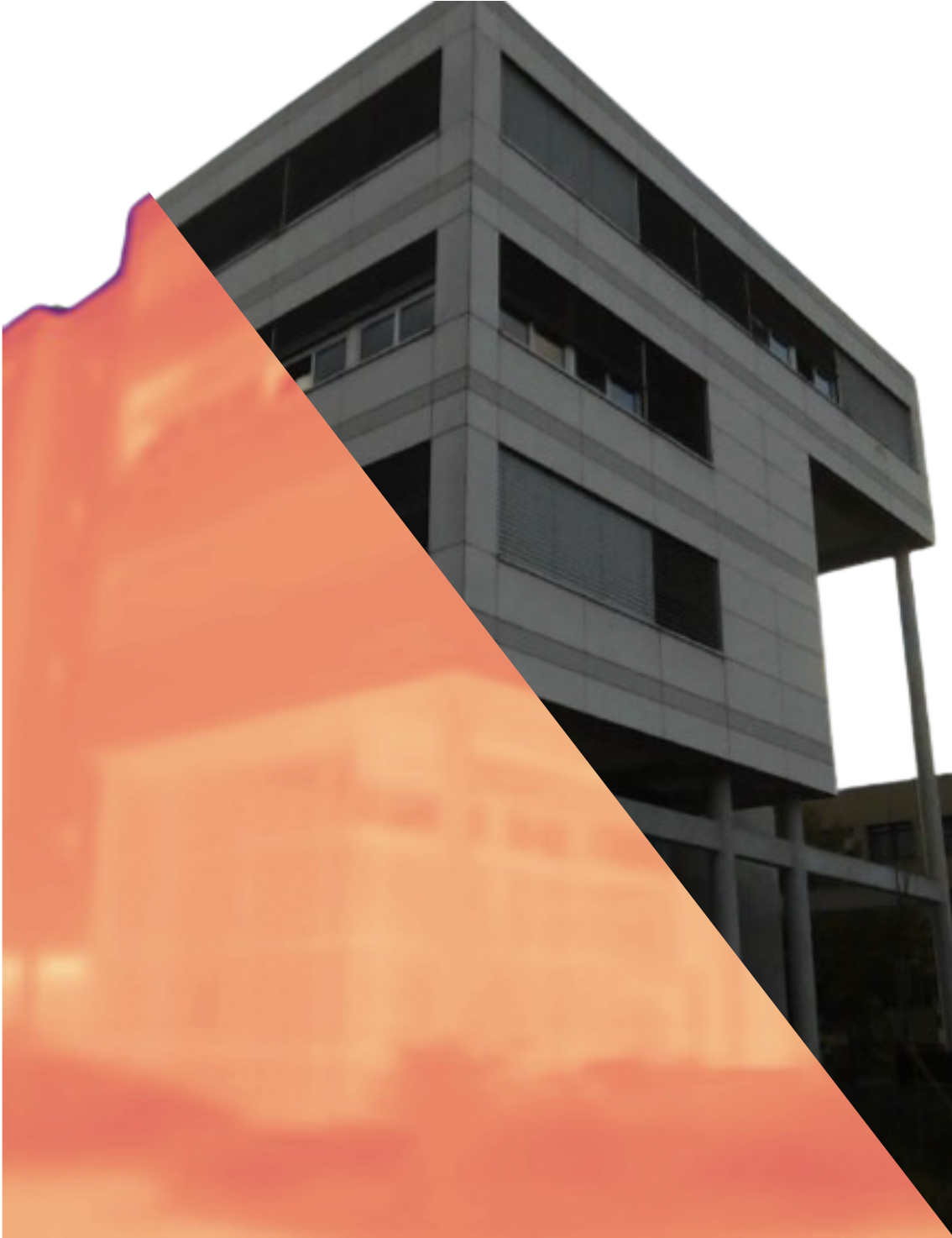}};

        \node[inner sep=0pt] (left) at (-1.8, -1.5)
        {\includegraphics[width=.07\textwidth]{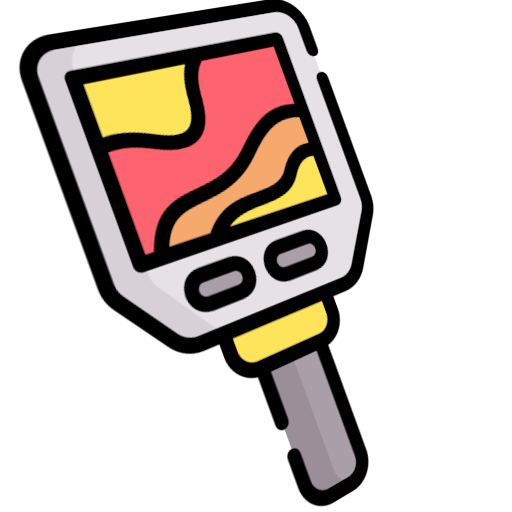}};
        \node[inner sep=0pt] (right) at (1.8,-1.5)
        {\includegraphics[width=.07\textwidth]{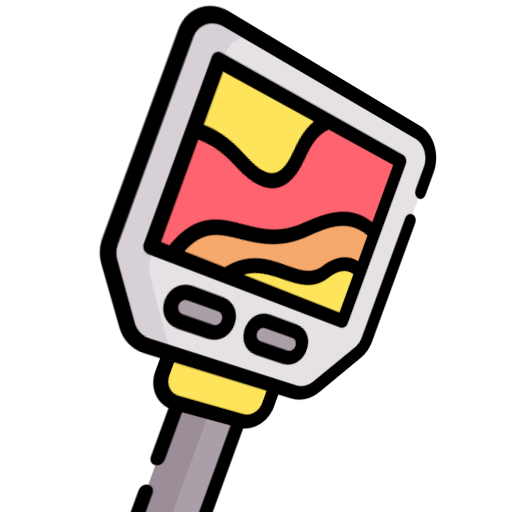}};
        \node[inner sep=0pt] (center) at (0,-2.5)
        {\includegraphics[width=.07\textwidth]{images/thermal-imaging2.png}};

        \draw[dblarw={white}{2pt}{1pt}] (left) -- (-0.5, -0.5);
        \draw[dblarw={white}{2pt}{1pt}] (right) -- (0.5, -0.5);
        \draw[dblarw={white}{2pt}{1pt}] (center) -- (0, -0.75);

        \node[align=center, below=0.1cm of left] {Thermal/RGB\\Image 1};
        \node[align=center, below=0.1cm of right] {Thermal/RGB\\Image 2};
        \node[align=center, below=0.1cm of center] {Thermal/RGB\\Image 3};
    \end{scope}

    \begin{pgfonlayer}{background}
        \node[category box, fill=backgroundColor!5, draw=backgroundColor, fit=(training_set)] {};
    \end{pgfonlayer}

    \begin{scope}[local bounding box=applications_set]
        \node[align=center, below=9.25cm of title, font=\bfseries, text width=7.2cm] (description) {Render thermal and RGB views};

        \node[align=left, below=0.1cm of description, text width=7.2cm] (applications) {
            \noindent Applications:
            \begin{itemize}[itemsep=0pt, leftmargin=0.15cm]
                \item Building retrofit
                \item Thermal leakage detection
                \item Infrastructure inspection
            \end{itemize}
        };

    \end{scope}

    \begin{pgfonlayer}{background}
        \node[category box, fill=backgroundColor!5, draw=backgroundColor, fit=(applications_set)] {};
    \end{pgfonlayer}

    \draw[ultra thick, -latex] (0, -4.2) -- (0, -5.65);


    \node[align=center, font=\bfseries, right=5cm of title] (thermoscene) {ThermoScenes};

    \begin{scope}[local bounding box=dataset_stats]
        \node[align=center, below=0.2cm of thermoscene, font=\bfseries, text width=7.2cm] (stats) {Dataset statistics};

        \node[align=left, below=-0.5cm of stats, text width=7.2cm] (statsdata) {
            \begin{itemize}[itemsep=0pt, leftmargin=0.15cm]
                \item Temperature range from -16.2$^\circ$C to 87.3$^\circ$C
                \item Eight building facade scenes
            \end{itemize}
        };

        \foreach \i/\x/\y/\thermal/\rgb in { 1/4.6/-1.1/{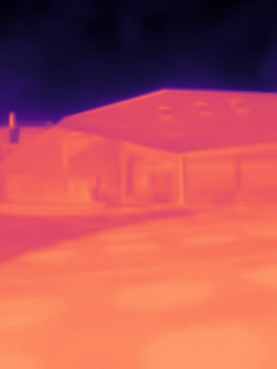}/{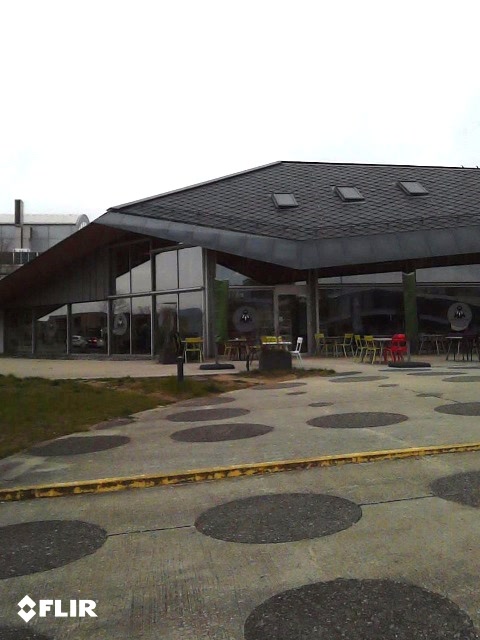}, 2/6.3/-1.1/{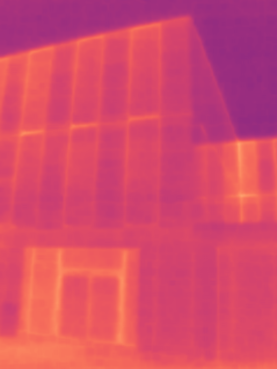}/{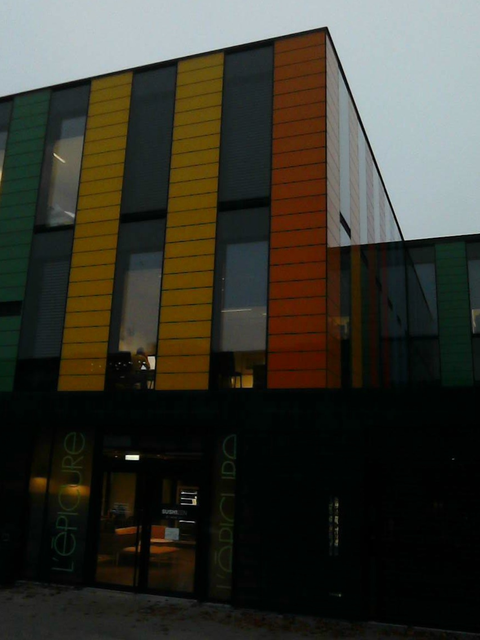}, 3/8/-1.1/{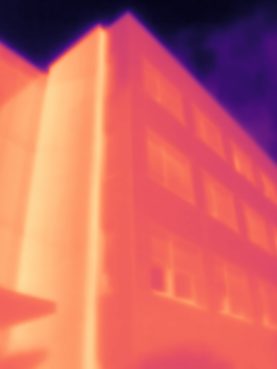}/{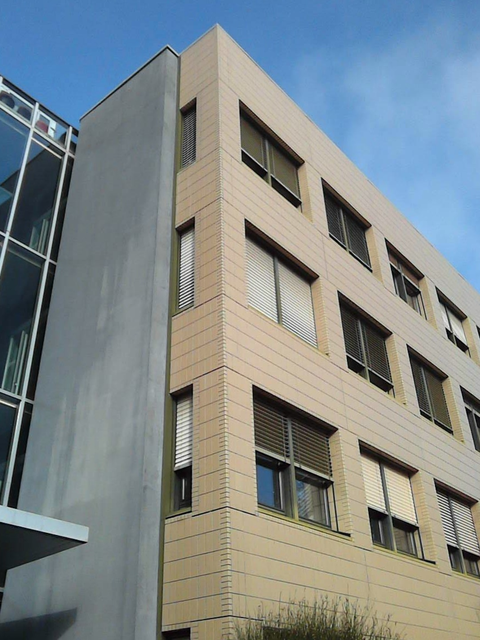},
        4/9.7/-1.1/{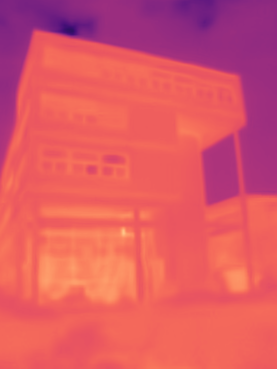}/{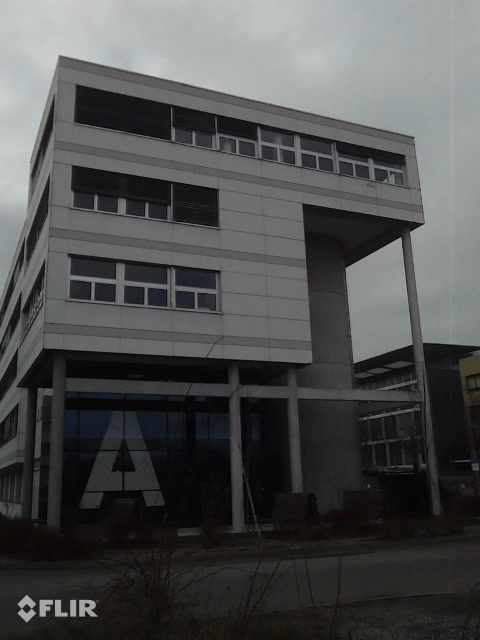}}
        {
        \node[inner sep=0pt] (image\i) at (\x, \y)
        {\includegraphics[width=.125\textwidth]{\thermal}};

        \begin{scope}
            \clip (image\i.north west) -- (image\i.south east) -- (image\i.north east) -- cycle;
            \node[inner sep=0pt] at (\x, \y)
            {\includegraphics[width=.125\textwidth]{\thermal}};
        \end{scope}

        \begin{scope}
            \clip (image\i.north west) -- (image\i.south east) -- (image\i.south west) -- cycle;
            \node[inner sep=0pt] at (\x, \y)
            {\includegraphics[width=.125\textwidth]{\rgb}};
        \end{scope}
        }

        \foreach \i/\x/\y/\thermal/\rgb in {
        5/4.6/-3.2/{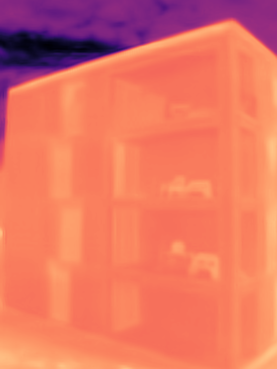}/{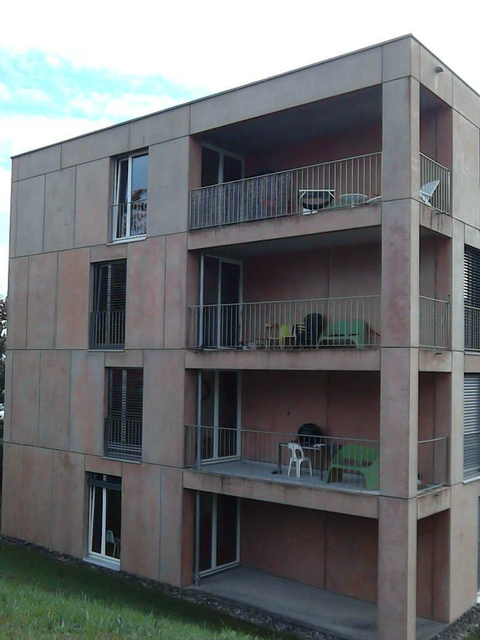},
        6/6.3/-3.2/{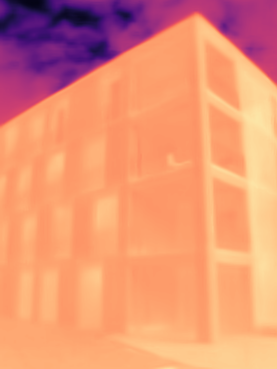}/{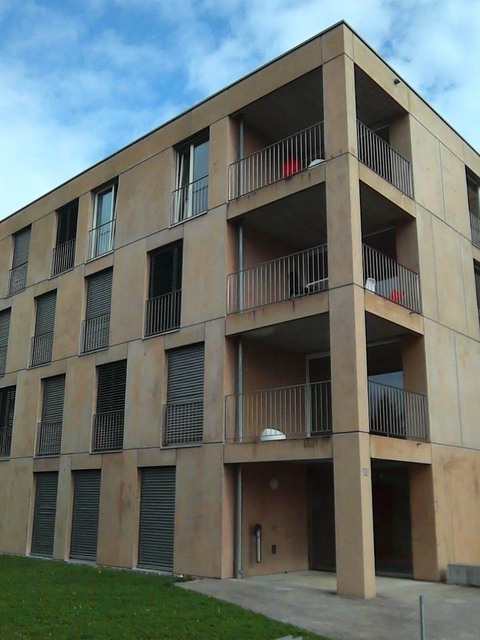},
        7/8/-3.2/{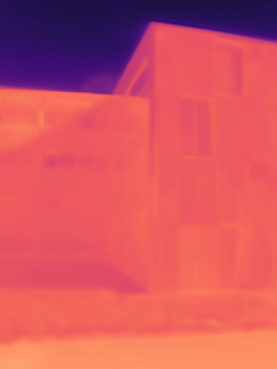}/{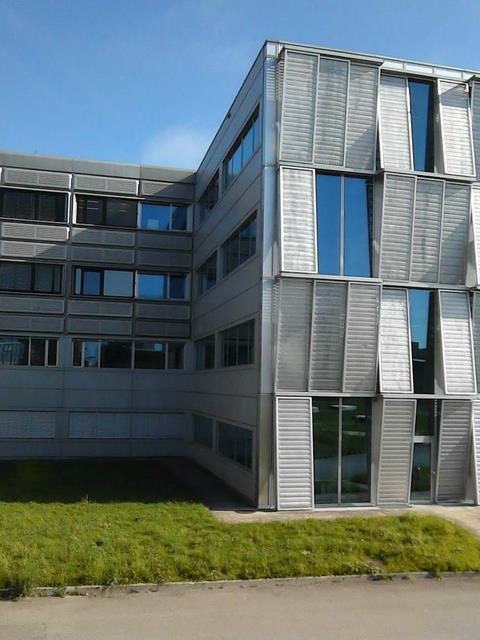},
        8/9.7/-3.2/{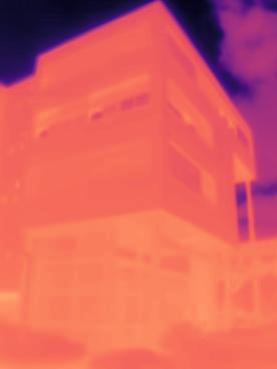}/{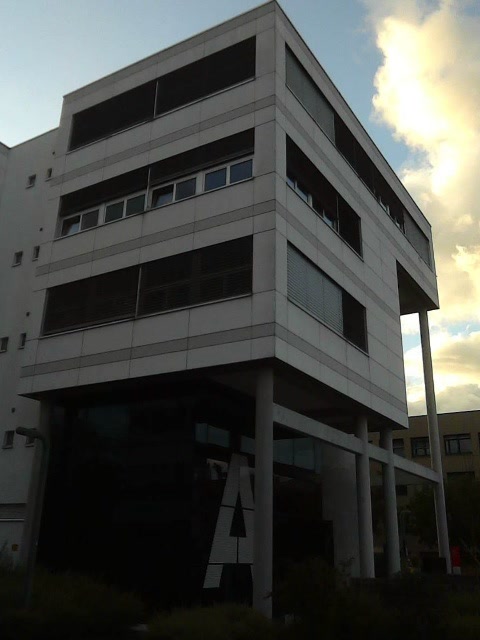}}
        {
        \node[inner sep=0pt] (image\i) at (\x, \y)
        {\includegraphics[width=.125\textwidth]{\thermal}};

        \begin{scope}
            \clip (image\i.north west) -- (image\i.south east) -- (image\i.north east) -- cycle;
            \node[inner sep=0pt] at (\x, \y)
            {\includegraphics[width=.125\textwidth]{\thermal}};
        \end{scope}

        \begin{scope}
            \clip (image\i.north west) -- (image\i.south east) -- (image\i.south west) -- cycle;
            \node[inner sep=0pt] at (\x, \y)
            {\includegraphics[width=.125\textwidth]{\rgb}};
        \end{scope}
        }

        \node[align=left, below=4.2cm of statsdata, text width=7.2cm] (statsdata2) {
            \begin{itemize}[itemsep=0pt, leftmargin=0.15cm]
                \item Eight other scenes
            \end{itemize}
        };

        \foreach \i/\x/\y/\thermal/\rgb in {
        9/4.6/-5.8/{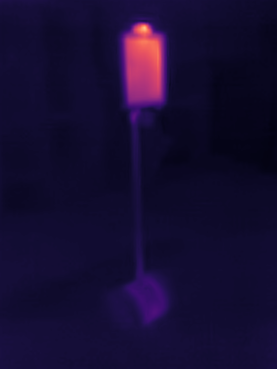}/{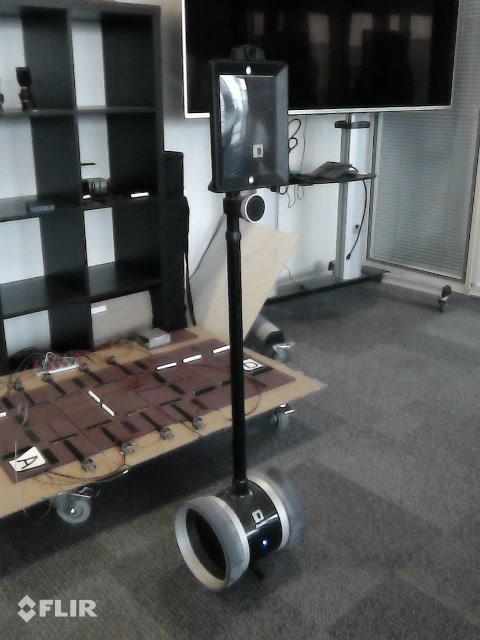},
        10/6.3/-5.8/{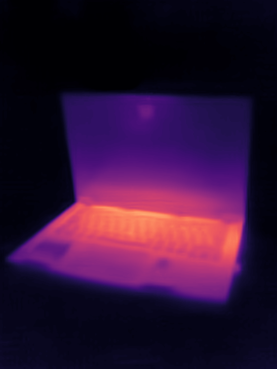}/{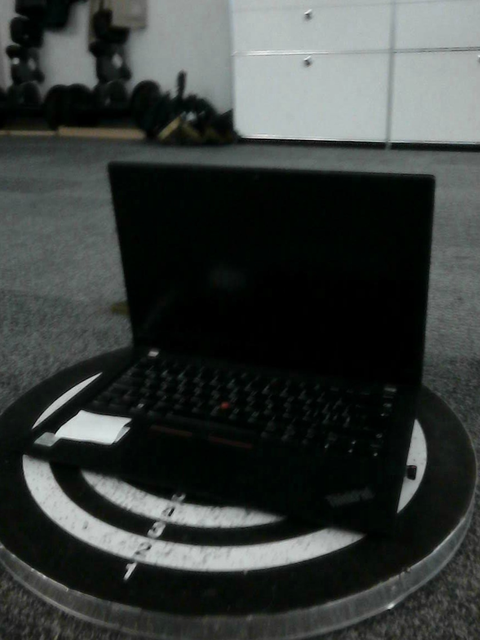},
        11/8/-5.8/{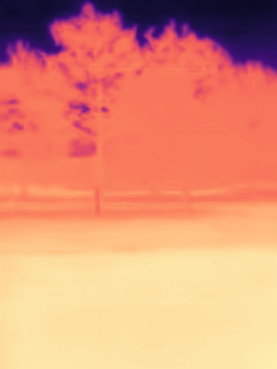}/{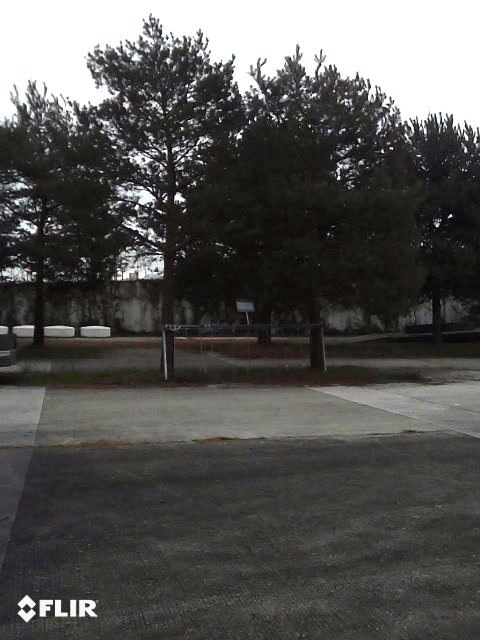},
        12/9.7/-5.8/{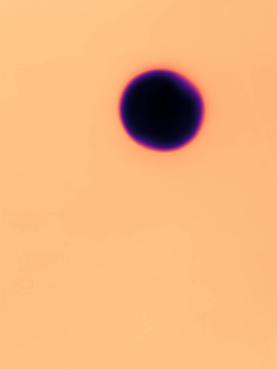}/{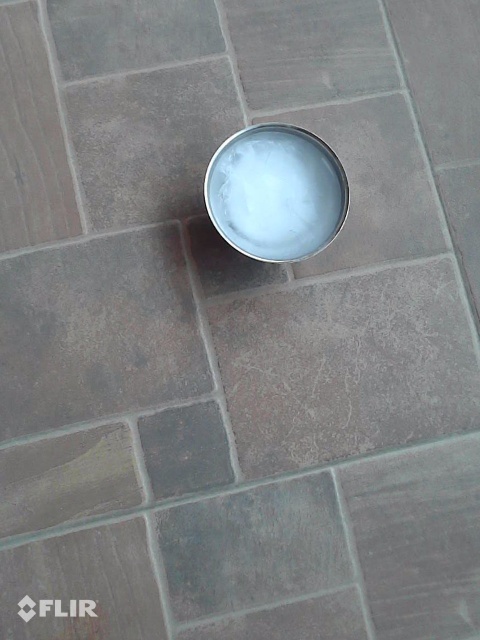}}
        {
        \node[inner sep=0pt] (image\i) at (\x, \y)
        {\includegraphics[width=.125\textwidth]{\thermal}};

        \begin{scope}
            \clip (image\i.north west) -- (image\i.south east) -- (image\i.north east) -- cycle;
            \node[inner sep=0pt] at (\x, \y)
            {\includegraphics[width=.125\textwidth]{\thermal}};
        \end{scope}

        \begin{scope}
            \clip (image\i.north west) -- (image\i.south east) -- (image\i.south west) -- cycle;
            \node[inner sep=0pt] at (\x, \y)
            {\includegraphics[width=.125\textwidth]{\rgb}};
        \end{scope}
        }

        \foreach \i/\x/\y/\thermal/\rgb in {
        13/4.6/-7.9/{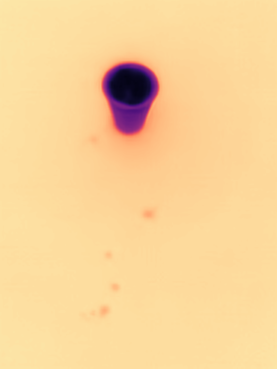}/{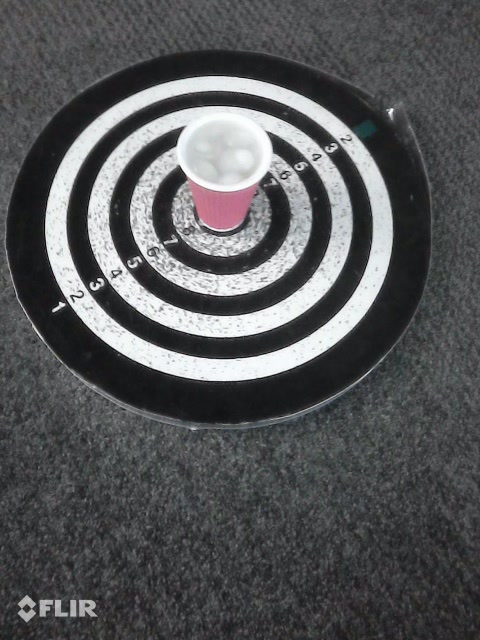},
        14/6.3/-7.9/{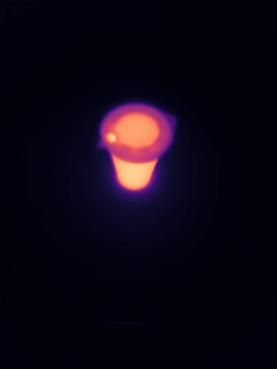}/{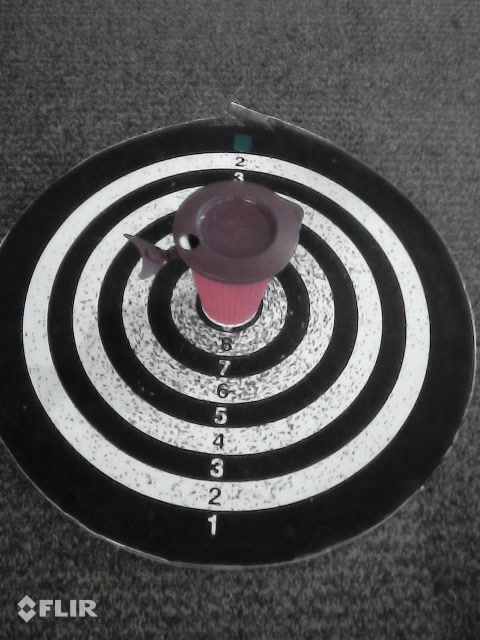},
        15/8/-7.9/{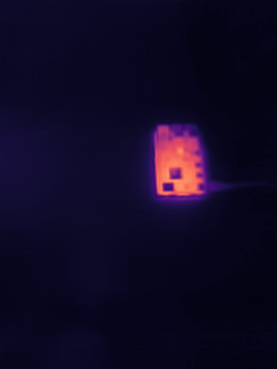}/{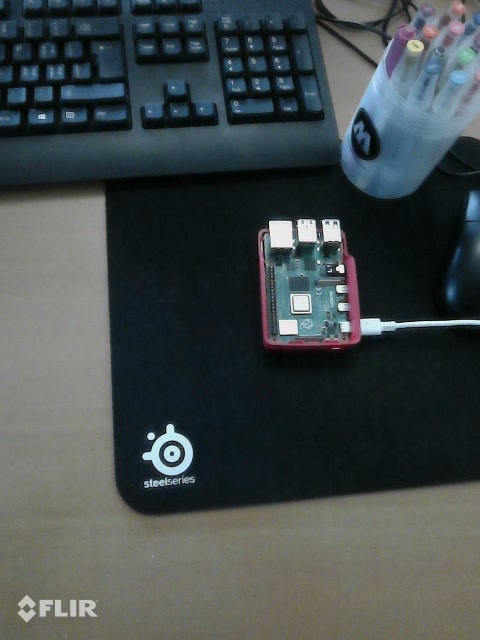},
        16/9.7/-7.9/{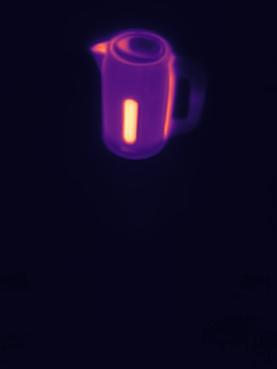}/{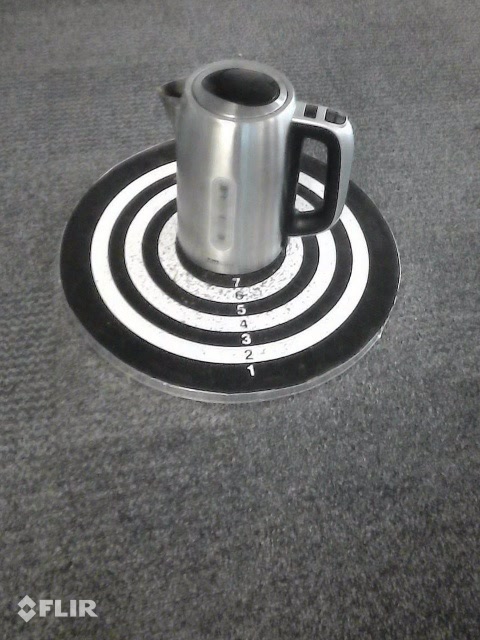}}
        {
        \node[inner sep=0pt] (image\i) at (\x, \y)
        {\includegraphics[width=.125\textwidth]{\thermal}};

        \begin{scope}
            \clip (image\i.north west) -- (image\i.south east) -- (image\i.north east) -- cycle;
            \node[inner sep=0pt] at (\x, \y)
            {\includegraphics[width=.125\textwidth]{\thermal}};
        \end{scope}

        \begin{scope}
            \clip (image\i.north west) -- (image\i.south east) -- (image\i.south west) -- cycle;
            \node[inner sep=0pt] at (\x, \y)
            {\includegraphics[width=.125\textwidth]{\rgb}};
        \end{scope}
        }
    \end{scope}

    \begin{pgfonlayer}{background}
        \node[category box, fill=backgroundColor!5, draw=backgroundColor, fit=(dataset_stats)] {};
    \end{pgfonlayer}

\end{tikzpicture}

%% file: introduction.tex
\section{Introduction} \label{sec:intro}

\begin{figure}[t]
  \centering
  \input{images/summary2.tikz}
  \caption{
    Overview of the capabilities of the proposed ThermoNeRF. It is a multimodal NeRF-based approach using paired thermal and RGB images. ThermoNeRF demonstrates enhanced geometry and thermal information estimation compared to thermal only models which cannot recover the geometry, or RGB+Thermal NeRF models for which information from the RGB image leaks into the temperature of the final rendered view.
  }

  \vspace{-3mm}
  \label{fig:thermonerf}
\end{figure}
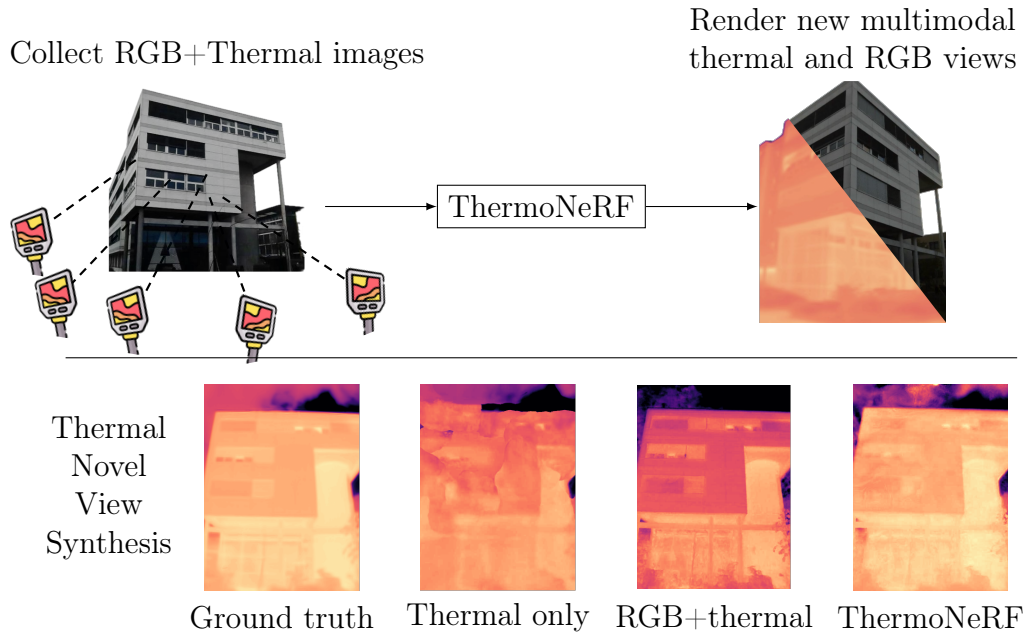

Thermal scene reconstruction offers significant potential for a wide range of applications, including building energy consumption analysis~\cite{HAM2013395}, non-destructive testing~\cite{article_thermal_imaging}, infrastructure inspection and monitoring~\cite{article_thermal_imaging, buildings13112829}, and optimization of the cost-effectiveness of building facade retrofit~\cite{ham2015three}---notably, building facades play a crucial role in regulating the energy flowing in and out of buildings~\cite{Nardi_2018}.
However, while thermal cameras capture radiation emitted by objects as a function of temperature, thermal images inherently  lack texture and exhibit  low edge contrast due to scattering and reflection from various surfaces, resulting in an effect known as ghosting~\cite{bao2023thermal}.
Consequently, thermal images are often not considered during geometric reconstruction, with 3D thermal models  predominantly  constructed using photogrammetry with large amounts of RGB images~\cite{maset_photogrammetric_2017, de2022generation, MARTIN2022112540}. In these models, temperature information is merely  projected  onto the reconstructed geometry 
post-reconstruction~\cite{RAMON2022112425}, leading to  discrepancies between the geometry and the temperature data.
The development of thermal scene reconstruction algorithms also faces challenges due to the lack of benchmark datasets specifically designed for this task, especially in the context of building facade reconstruction.

Recently, Neural Radiance Fields (NeRFs)~\cite{mildenhall2020nerf} have achieved great success in 3D reconstruction and novel view synthesis
by learning  an implicit representation of a scene from a set of sparse RGB images.
While most NeRF models focus on learning from RGB images, they have also been extended to other sensor modalities such as hyperspectral images~\cite{poggi_cross-spectral_2022}, acoustic signals~\cite{chen2023novel}, point clouds~\cite{zhu_multimodal_2023, zhang2023nerf}, or thermal images~\cite{lin2024thermalnerf,ozer2024exploring}.
However, 
prior research on NeRF utilizing thermal images has either relied on high-resolution thermal images captured with costly equipment or emphasized the visual quality of the reconstruction in either low-light conditions~\cite{xu2024leveraging} or against occlusions~\cite{lin2024thermalnerf}, without prioritizing the precision of the estimated temperature which hinders their practical use in different applications.

To bridge  these research gaps, we propose (1) ThermoNeRF (\underline{Thermo}graphic \underline{NeRF}), a multimodal NeRF model capable of rendering unseen views in both RGB and thermal modalities with a focus on temperature rendering quality and not only image quality (see \cref{fig:thermonerf}) and (2) ThermoScenes, the first paired RGB+thermal dataset comprising sixteen diverse scenes---eight building facade scenes and eight everyday objects scenes.
The design of ThermoNeRF is guided by the specific properties of thermal images.
ThermoNeRF uses a shared density MLP that leverages the visual features of the RGB modality to learn the geometry of the scene while ensuring consistency with the thermal measurements.
On the other hand, color and thermal information are learned through decoupled MLPs, preventing the influence of colors on the estimated temperatures, and vice-versa.
Both the dataset\footnote{https://doi.org/10.5281/zenodo.10835108} and the code\footnote{https://github.com/Schindler-EPFL-Lab/thermo-nerf} (including instructions to submit new scenes to the dataset) are available online.

Our contributions are summarized as follows:

\begin{itemize}
  \item We propose ThermoNeRF, a multimodal NeRF capable of rendering both thermal and RGB views jointly and conduct extensive experiments to validate how ThermoNeRF's architecture allows the RGB modality to guide the density estimation for thermal reconstruction while preventing information leakage between both modalities, hence leading to accurate temperature estimations.
  \item We provide ThermoScenes, the first RGB+thermal image dataset for 3D scene reconstruction and novel view synthesis, featuring 16 scenes with diverse temperature ranges and objects, including 8 building facades and 8 everyday object scenes.
  \item Finally, we present a comprehensive evaluation covering both  temperature estimation and reconstruction quality on unseen poses.

\end{itemize}

Our results demonstrate improved temperature estimation with no loss in reconstruction fidelity when compared to models trained using only thermal images or concatenated RBG+thermal images as input.

%% file: images/summary2.tikz
\begin{tikzpicture}[scale=1.15]

    \node[align=center] (input) at (-1.25, 0.5) {Collect RGB+Thermal images};

    \node[inner sep=0pt] (original) at (-1.5, -1.5)
    {\includegraphics[width=.4\textwidth]{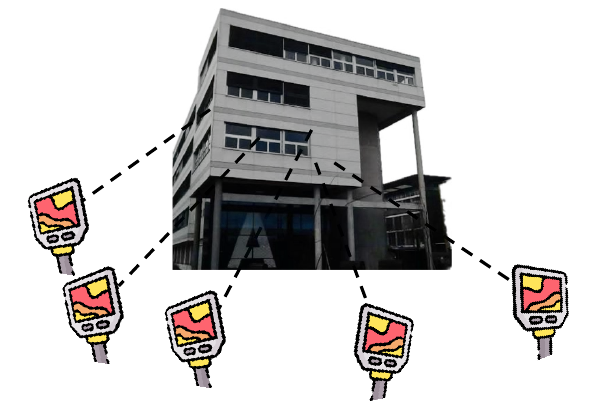}};




    \node[draw, align=center] (thermonerf) at (2.5,-1.25) {ThermoNeRF};
    \draw[-latex] (0, -1.25) -- (thermonerf);


    \node[inner sep=0pt] (merged) at (6.1,-1.2)
    {\includegraphics[width=.18\textwidth]{images/merged_spring_building.png}};

    \node[align=center] (reconstruction) at (6.1, 0.7) {Render new multimodal\\thermal and RGB views};
    \draw[-latex] (thermonerf) -- (5, -1.25);

    \draw[] (-3, -3) -- (8, -3);

    \node[align=center] (results) at (-2.5,-4.5) {Thermal \\ Novel \\ View \\ Synthesis};
    
     \node[inner sep=0pt] (up) at (-0.5,-4.5){\includegraphics[width=.15\textwidth]{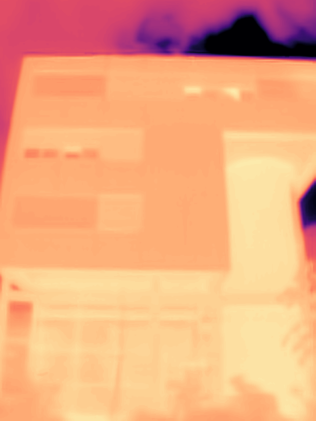}};
    \node[align=center] (results) at (-0.5,-6) {Ground truth};
    
    \node[inner sep=0pt] (up) at (2,-4.5){\includegraphics[width=.15\textwidth]{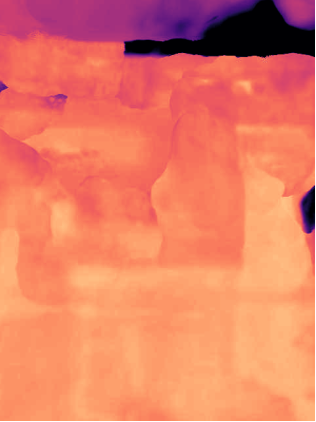}};
  
    \node[align=center] (results) at (2,-6) {Thermal only};
    
    \node[inner sep=0pt] (up) at (4.5,-4.5){\includegraphics[width=.15\textwidth]{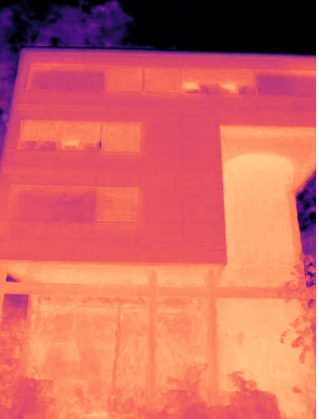}};
    
    \node[align=center] (results) at (4.5,-6) {RGB+thermal};
    
    \node[inner sep=0pt] (up) at (7,-4.5){\includegraphics[width=.15\textwidth]{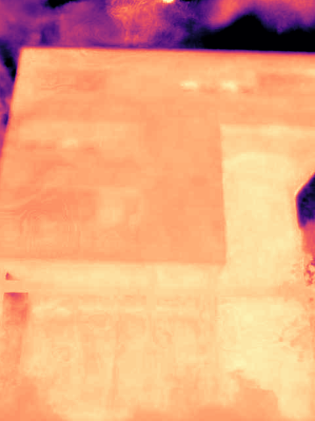}};
    \node[align=center] (results) at (7,-6) {ThermoNeRF};

\end{tikzpicture}

%% file: related_work.tex
\section{Related Work} \label{sec:background}

\subsection{Thermal Computer Vision}
\label{sec:rw:tcompvis}

Thermal cameras measure thermal radiation emitted by any matter with a temperature above absolute zero Kelvin.
While most practical temperature ranges fall within the infrared (IR) region of the electromagnetic spectrum~\cite{Hu2019}, thermal cameras operate in the mid- to far-IR spectrum, unlike near-infrared (NIR) sensors, which work closer to the visible range.
Thermal cameras detect thermal radiations that are scattered and reflected by objects and the environment, resulting in thermal images that are naturally textureless and soft---a phenomenon known as the ghosting effect \cite{bao2023thermal}.

While researchers have explored object detection~\cite{devaguptapu2019borrow} and tracking~\cite{berg2015thermal} using infrared sensors, as well as the analysis of thermal images~\cite{janssens2017deep, luo2019temporal, liu2020deep}, due to their soft and textureless nature, working with thermal images is a challenge \cite{bao2023thermal}.
Hence, early studies modeling thermal data 
predominantly relied on range measurements---e.g. LiDAR~\cite{laguela_energy_2011}, depth cameras~\cite{VIDAS2013445, schramm_combining_2022} or both~\cite{campione_3d_2020}---to construct 3D models, onto which thermal information was subsequently projected.
However, those approaches require a time-consuming and error-prone manual calibration of sensor transformations, their precision depends on the accuracy of mapping algorithms, and in the case of depth cameras, are limited by their range.
To remove the dependence on range measurements, \citet{maset_photogrammetric_2017, de2022generation} use photogrammetry to construct 3D models from RGB images.
While these methods achieve high model accuracy, they depend on a dense set of RGB and thermal images.
On the other hand, \citet{chen_3d_2015} and \citet{sentenac_automated_2018} used thermal images for 3D reconstruction and temperature estimation but their results are limited to thermal images with high contrast and do not generalize to large structures like building facades.
Hence, building thermal envelope analysis is usually done by mapping thermal data on existing models~\cite{NATEPHRA2017194}.

\subsection{NeRF and Multimodality}

Traditional methods in 3D reconstruction, such as Structure from Motion~\cite{Schonberger_2016_CVPR} or Multi-View Stereo~\cite{schonberger2016pixelwise}, depend on dense input data and scenes rich in geometric features.
The seminal work by \citet{mildenhall2020nerf} introduced NeRF, enabling the learning of implicit representations of 3D scenes from sparse sets of RGB images for novel view synthesis.
Subsequent research efforts have extended and refined the NeRF framework, addressing challenges related to scalability~\cite{martin2021nerf}, level of detail~\cite{barron2021mipnerf, barron2023zipnerf} and efficiency~\cite{yu2021plenoctrees}.

To enhance scene reconstruction, the first multimodal approach incorporated depth data to guide the learning process~\cite{Deng2021DepthsupervisedNF}.
However, recent studies have demonstrated that integrating other modalities---e.g. RGB 2D images and LiDAR 3D point clouds~\cite{zhu_multimodal_2023} or spectral information from filtered RGB inputs~\cite{li_spec-nerf_2023}---into NeRF can also yield improved scene representations.

Closely related to our work on thermal imaging, \citet{poggi_cross-spectral_2022} integrates near-infrared and RGB images to create a composite NIR-RGB 3D scene. 
However, NIR images share similar features with RGB images, as they are closer to the visible spectrum compared to thermal images.

Unlike existing methods that rely on multimodal data with salient visual features, ThermoNeRF combines RGB and thermal features to learn consistent scene geometries across both sensor modalities, while avoiding contamination from each modality in the rendered views.

Concurrently to our work, other studies have investigated the use of NeRFs and thermal imaging~\cite{lin2024thermalnerf, ozer2024exploring, ye2024thermal}.
However, they either rely on high-quality thermal images or focus on adverse conditions---such as low-light environments~\cite{xu2024leveraging}---to enhance the visual quality of generated images, typically evaluated using RGB-specific metrics such as the PSNR, SSIM, and LPIPS. For instance, \citet{xu2024leveraging} only evaluate visible light image synthesis, neglecting the quality of the resulting thermal images.
In contrast, our work prioritizes the accuracy of predicted temperatures in the synthesized novel views, which is crucial for practical, real-world applications such as thermal modeling of buildings.

\subsection{Thermal 3D Reconstruction for Building Retrofit}

Both 3D reconstruction and thermal imaging have been employed for building evaluation and retrofit.
However, most state-of-the-art methods require either expensive and complex software~\cite{ham2015three}, extensive data-collection and cleaning efforts~\cite{Bulatov2024UsingPM}, or a combination of both~\cite{Ramani2023ADU, Evangelisti2015EnergyRS}.

Most similar to our work, \citet{Ham20143DVO} uses photogrammetry to reconstruct four indoor scenes and detect condensation issues and thermal resistance of the wall.
However, their approach shares the same limitations as the photogrammetry-based methods discussed in \cref{sec:rw:tcompvis}.

In conclusion, while both 3D reconstruction and thermal imaging have been employed for building evaluation and retrofit purposes, current state-of-the-art methods often present significant complexities and requirements.
Our research tackle those issues by presenting a more practical and accessible approach to thermal modeling of buildings.
ThermoNeRF combines the use of sparse data as input (i.e., flexibility in the data collection) and joint multi-modal reconstruction that does not rely on thermal to RGB projection.

%% file: method.tex
\section{Preliminary}
\label{sec:preliminary}

NeRF~\cite{mildenhall2020nerf} learns an implicit scene representation as a continuous 5D vector-valued function, mapping a 3D position $\text{x} = (x,y,z)$ and a viewing direction $\text{d} = (\theta, \phi)$ to a color $c$ and a volume density $\sigma$. This mapping is applied  along rays projected in space from each pixel of an image captured  by a camera at pose $(\text{x}, \text{d})$.
NeRF comprises  two Multi-Layer Perceptron (MLP) networks: one predicts the density based solely on the 3D location, and the other predicts color based on the 3D location and the viewing direction, making the color prediction view-dependent to account for non-Lambertian effects on surfaces. Finally, to reconstruct a view, rays are sampled through the scene, and the color of each ray is rendered using volume rendering techniques.

Nerfacto~\cite{Tancik_2023} is a recent extension of NeRF that integrates components from several state-of-the-art models, such as MIP-NeRF~\cite{barron2021mipnerf}, MIP-NeRF 360 \cite{barron2022mipnerf360}, Instant-NGP~\cite{mueller2022instant} and NeRF-W~\cite{martin2021nerf}, amongst others.
The model's inputs include an additional appearance embedding, which is a per-image embedding that accounts for differences in exposure and lighting conditions across different training views.
Furthermore, sinusoidal-based positional encoding is replaced  with hash encoding for positions and spherical harmonics for viewing direction. Hence, Nerfacto is defined as follows:
\begin{equation}
  \begin{split}
    [\sigma, f] & = \text{MLP}_{\text{dens}}\left(\phi_H(\text{x})\right), \\
    c & =\text{MLP}_{\text{rgb}}\left(f, \phi_{SH}(\text{d}), A\right),\\
  \end{split}
  \label{sec:method:eq:nerf}
\end{equation}
where $f$ are intermediate features, c is colors
, $\phi_H(\cdot)$ denotes  the hash encoding, $\phi_{SH}(\cdot)$ is the spherical harmonics encoding and $A$ represents the appearance embeddings.

Additionally, Nerfacto uses pose refinement to further optimise the poses predicted for every view.
It uses a piecewise sampler to balance sampling through the fine details of the scene and the distant views, and a proposal sampler that emphasizes the density regions of the scene which impact reconstruction quality the most.
ThermoNeRF extends Nerfacto to allow reconstruction as well as novel view synthesis of thermal and RGB images.

\section{ThermoNeRF}
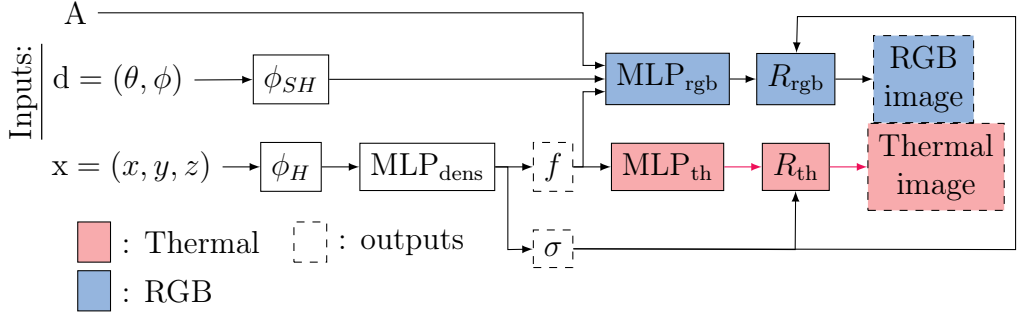
\begin{figure}[t]
  \centering
  \input{images/thermo_nerf.tikz}
  \caption{ThermoNerf architecture; in red are parts of the network related to the generation of thermal images, while blue represents the parts related to the generation of RGB images.
  The $\text{MLP}_{\text{th}}$ is only dependent on the intermediate features $f$ as input. A is appearance embeddings that accounts for difference in exposure for RGB images.
  }
  \vspace{-3mm}
  \label{fig:thermo-nerf}
\end{figure}

In this section, we present ThermoNeRF, a NeRF model capable of learning an implicit scene representation in both thermal and RGB, from a sparse set of RGB and thermal images.
To synthesize realistic and accurate RGB and thermal novel views, ThermoNeRF retains geometric features captured by the RGB modality, while ensuring that temperature estimates are independent of scene color variations and details, and that they have a unique value for a given 3D point.
Thus, our design choices are guided by two properties of thermal images: 1) thermal images are inherently \emph{soft and textureless}, 2) surface temperatures are \emph{independent of the viewing direction}.

Refer to \cref{fig:thermo-nerf} for a flowchart of the method;
implementation of the model and evaluations can be found online\footnote{\url{https://github.com/Schindler-EPFL-Lab/thermo-nerf}}.

\subsection{Thermal Image Rendering}

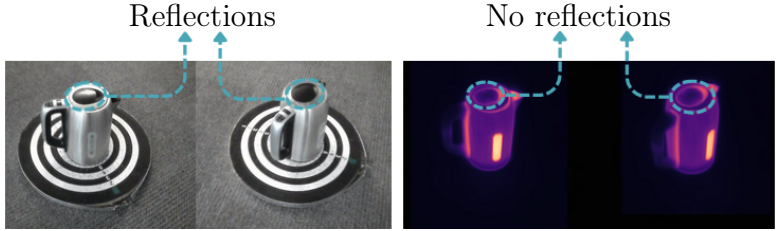
\begin{figure}[t]
  \centering
  \input{images/lambertian.tikz}
  \caption{
    Non-Lambertian effects---i.e. light reflections---present in the RGB images (left) depend on the angle of view and are not present in thermal images (right).
    Furthermore, textures and edge features in the thermal images are soft due to the ghosting effect, as opposed to the sharpness of the RGB image and its background.
  }
  \label{fig:lighting_effect}
  \vspace{-3mm}
\end{figure}

As seen in the thermal images of \cref{fig:lighting_effect}, the ghosting effect makes thermal images soft and textureless.
Consequently, while textures and sharp geometric features in RGB images allow NeRF models to accurately estimate scene density, NeRF struggles to learn densities solely from thermal images---see \cref{fig:thermal-only} for experimental results of training Nerfacto on thermal images only.
Therefore, a single density $\text{MLP}_{\text{dens}}$ is shared by the modalities, enabling the network to construct a geometric representation informed by RGB information while ensuring geometric consistency with the thermal modality.
This common density network is defined in the same way as in \cref{sec:preliminary}:
\begin{equation}
  \label{rgbt_nerf}
  [\sigma, f] = \text{MLP}_{\text{dens}}(\phi_H(\text{x}))
\end{equation}

Since the aim of ThermoNeRF is to synthesize images that faithfully represent temperature at the surface, details present in the RGB images (such as edges, contrast, texture) should not ``leak'' into the synthetic thermal views, and colors in the RGB image should not influence the value of the synthesized temperatures in thermal images.
Therefore, we propose to use separate MLPs to estimate each modality: $\text{MLP}_{\text{th}}$ predicts temperatures along each ray, while $\text{MLP}_{\text{rgb}}$ predicts RGB values.


Furthermore, as depicted in \cref{fig:lighting_effect}, color information is view-dependent---e.g., due to non-Lambertian effects---which is why, as seen in \cref{sec:method:eq:nerf}, $\text{MLP}_{rgb}$ takes as input $A$, and $\phi_{SH}(\text{d})$ to account for view direction.
Thermal images are also view-dependent, not due to non-Lambertian effects but due to thermal reflections of nearby heat sources---for example, ceramic, metal, and glass are excellent thermal reflectors.
However, in the context of non-destructive testing of infrastructure and infrastructure modeling, it is more interesting to reconstruct an image of the ``real'' temperature at the surface, rather than a realistic reconstruction of the scene influenced by thermal reflections, i.e. temperature should be invariant with respect to the viewing direction.
Thus, $\text{MLP}_{\text{th}}$ only takes the intermediate features $f$ of $\text{MLP}_{\text{dens}}$ as input, disregarding $A$, and $\phi_{SH}$.

Formally, ThermoNeRF is defined as follows:

\begin{equation}
  \label{rgbt_nerf}
  \begin{aligned}
    [\sigma, f] & = \text{MLP}_{\text{dens}}(\phi_H(\text{x})),        \\
    c           & =\text{MLP}_{\text{rgb}}(f, \phi_{SH}(\text{d}), A), \\
    t           & = \text{MLP}_{\text{th}}(f),
  \end{aligned}
\end{equation}

where $t$ denotes the predicted temperature values and A is the appearance embedding.

\subsection{Loss Functions}

We use separate reconstruction loss functions for each modality. The final reconstruction loss $\mathcal{L}$ is expressed as the sum of the Mean Square Error (MSE) losses of the RGB and thermal outputs, respectively denoted $\mathcal{L}_{\text{rgb}}$ and $\mathcal{L}_{\text{th}}$. Similar to Nerfacto, we add the interlevel and distortion losses---respectively $\mathcal{L}_{\text{interl.}}$ and $\mathcal{L}_{\text{dist}}$---which were initially used in MIP-NeRF 360 \cite{barron2022mipnerf360}, to optimize the proposal sampler and to reduce distortions respectively. Therefore, our final loss is defined as follows:
\begin{equation}
  \mathcal{L}_{\text{ThermoNeRF}} = \lambda_r \mathcal{L}_{\text{rgb}} + \lambda_t \mathcal{L}_{\text{th}} + \mathcal{L}_{\text{dist}} + \mathcal{L}_{\text{interl.}}
  \label{eq:total_loss}
\end{equation}

While we tried different values for $\lambda_r$ and $\lambda_t$, we empirically found $\lambda_r =\lambda_t =1$ to be the best value and will thus use it in the rest of this paper.

%% file: images/thermo_nerf.tikz
\usetikzlibrary{positioning}

\begin{tikzpicture}[node distance = 0.5cm, every node/.style={minimum height=0.5cm}, scale=1]
    \node[align=center, rotate=90] (input) at (-1.4, 1) {\underline{Inputs:}};

    \node[] (xyz) at (0, 0) {$\text{x} = (x, y, z)$};
    \node[draw, fill=OrangeRed!40] (placeholder) at (-0.5, -1)  {\phantom{t}};
    \node[right=0cm of placeholder] (legend) {: Thermal};
    
    \node[draw, right=0.3cm of legend, dashed] (placeholder2)  {\phantom{t}};
    \node[right=0cm of placeholder2] (legend2) {: outputs};
    
    \node[draw, below=0.1cm of placeholder, fill=RoyalBlue!40] (placeholder3)  {\phantom{t}};
    \node[right=0cm of placeholder3] (legend) {: RGB};
    
    \node[draw, right=of xyz] (g_x) {$\phi_H$};
    \node[draw, right=of g_x] (td)  {$\text{MLP}_{\text{dens}}$};

    \node[draw, right=of td, dashed] (int_feat)  {$f$};

    \node[draw, right=of int_feat, fill=OrangeRed!40] (tt) {$\text{MLP}_{\text{th}}$};
    \node[draw, above=of tt, fill=RoyalBlue!40] (tc) {$\text{MLP}_{\text{rgb}}$};
    \node[draw, above=of g_x] (sh) {$\phi_{SH}$};
    \node[] (dir) at (-0.18, 1.15) {$\text{d} = (\theta, \phi)$};
    \node[align=center] (app_embded) at (-0.75, 2.05) {A};

    
    \node[draw, right=of tt, align=center, fill=OrangeRed!40] (RT) {$R_{\text{th}}$};
    \node[draw, above=of RT, align=center, fill=RoyalBlue!40] (RC) {$R_{\text{rgb}}$};

    \node[draw, below=of int_feat, dashed] (density) {$\sigma$};

    \node[draw, right=of RT, align=center, dashed, fill=OrangeRed!40] (thermal_img) {Thermal\\image};
    \node[draw, right=of RC, align=center, dashed, fill=RoyalBlue!40] (rgb_img) {RGB\\image};

    \draw[-latex] (xyz) -- (g_x);
    \draw[-latex] (g_x) -- (td);
    \draw[-latex] (td) -- (int_feat);
    
    \draw[-latex] (td) -- (5, 0) -- (5, -1.1) -- (density);
    
    \draw[-latex] (int_feat) -- (tt);
    \draw[-latex] (int_feat) -- (6, 0) -- (6, 1) --(6.3, 1);

    \draw[-latex] (dir) -- (sh);
    \draw[-latex] (sh) -- (tc);
    \draw[-latex] (app_embded) -- (6, 2.05) -- (6, 1.35) --(6.3, 1.35);

    \draw[-latex] (tc) -- (RC);

    \draw[draw=OrangeRed, -latex] (tt) -- (RT);

    \draw[-latex] (density) -- (11.75, -1.1) -- (11.75, 2) -- (8.85, 2) -- (RC);
    \draw[-latex] (density) -- (8.82, -1.1) -- (RT);
    
    \draw[draw=OrangeRed, -latex] (RT) -- (thermal_img);
    \draw[-latex] (RC) -- (rgb_img);

\end{tikzpicture}

%% file: images/lambertian.tikz
\begin{tikzpicture}
    
    \node[inner sep=0pt] (up) at (0, 0)
    {\includegraphics[width=.75\textwidth]{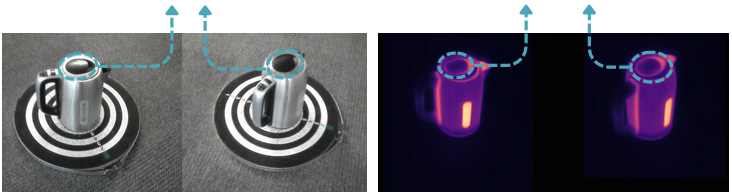}};

    \node[align=center] (input) at (-2.5, 1.5) {Reflections};
    \node[align=center] (input) at (2.5, 1.5) {No reflections};

\end{tikzpicture}

%% file: dataset.tex
\section{ThermoScenes Dataset}
\label{sec:dataset}

\input{images/dataset_table}

To evaluate RGB+thermal reconstruction and novel view synthesis methods, we have collected ThermoScenes, a new dataset comprising  paired RGB and thermal images for 16 scenes (eight building facades and eight other everyday scenes) with diverse temperature ranges.
A FLIR One Pro LT~\cite{flirone}---a dual RGB and thermal camera mounted on an iPhone and capable of simultaneously capturing aligned pairs of RGB and thermal images---was used to collect the data.
The FLIR One Pro LT thermal camera operates within a range of -20 °C to 120 °C, with a thermal accuracy of $\pm$ 3 °C.
In our study, evaluating the quality of 3D thermal reconstruction requires not only understanding the thermal accuracy but also the thermal precision of the camera.
We assess the precision to be approximately  $\pm 0.14^\circ$C.
To calculate the precision of our thermal camera, we fix the camera at a position and capture 20 consecutive images of a RaspberryPi connected to power for at least 10 minutes to ensure thermal stability. Our precision is then the average standard deviation of the temperatures of each pixel across the 20 images.


Furthermore, the FLIR software generates thermal images by overlaying thermal information onto RGB features---such as edges---using Multi-Spectral Dynamic Imaging (MSX) \cite{FLIR_msx}.
Instead, we provide and use raw thermal data extracted from MSX images using the FLIR Image Extractor library~\cite{PyPI}.

%
%

\cref{tbl:thermoscenes-summary} provides a summary of the scenes comprising the ThermoScenes dataset,
where each scene is illustrated by two example views in RGB and thermal respectively.
The test set for each scene 
is uniformly sampled across the collected camera poses.
The minimum and maximum measured temperatures are also reported.
It should be noted that in the Building (spring), Dorm 1, MED, and INR building scenes, the sensor outputs values outside of its operating range due to the variations of atmospheric radiation in the sky, where the readings of thermal cameras are subject to systematic inaccuracies~\cite{KRUCZEK2023128466}.
Given our focus on the object of interest---the building facade---we limit these values to the minimum of the operating range of the camera (-20°C) in our experiments.
However, for other applications where more precise temperature estimates for the sky might be needed, we provide the raw data to allow for potential post-processing (see \citet{KRUCZEK2023128466} and \citet{ZHAO2023110369}).



ThermoScenes is publicly available\footnote{\url{http://dx.doi.org/10.5281/zenodo.10835108}}
and we plan to gradually increase the size of the dataset.
A tutorial on how to collect, process, and train data on ThermoNeRF has also been released.\footnote{\url{https://github.com/Schindler-EPFL-Lab/thermo-nerf/blob/main/thermo\_scenes/docs/Collect\_new\_dataset.md}}

%% file: images/dataset_table.tex
\begin{table}[t]
    \caption{Summary of the collected ThermoScenes dataset showing example views with paired RGB and thermal images, the number of train/test views and the temperature range for each scene, for the eight building scenes (left) and the eight everyday scene (right).}
    \begin{tabular}{cc}%
        \resizebox{.48\textwidth}{!}{%
            \begin{tabular}{l cccc}
                \toprule
                Scene                                                                                                     & RGB                                                                                             & Thermal & \#views & \makecell{Temp. \\ range }\\
                \midrule

                \makecell[l]{Building                                                                                                                                                                                                                             \\ (Spring) } & \adjustbox{valign=c}{\includegraphics[width=.1\textwidth]{images/dataset/building-a-spring/rgb.jpeg}} &
                \adjustbox{valign=c}{\includegraphics[width=.1\textwidth]{images/dataset/building-a-spring/thermal.jpeg}} &
                \makecell{107 (train)                                                                                                                                                                                                                             \\ 15 (test)}&
                \makecell{-62.5°C*                                                                                                                                                                                                                                \\ 19.7°C }\\

                \makecell[l]{Exhibition                                                                                                                                                                                                                           \\ Building } & \adjustbox{valign=c}{\includegraphics[width=.1\textwidth]{images/dataset/pavilion/rgb.jpg}} &
                \adjustbox{valign=c}{\includegraphics[width=.1\textwidth]{images/dataset/pavilion/thermal.PNG}}           &
                \makecell{119 (train)                                                                                                                                                                                                                             \\ 16 (test)}&
                \makecell{-11.3°C                                                                                                                                                                                                                                 \\ 14.0°C }\\

                \makecell[l]{Building                                                                                                                                                                                                                             \\ (Winter) } & \adjustbox{valign=c}{\includegraphics[width=.1\textwidth]{images/dataset/building-a-winter/rgb.JPG}} &
                \adjustbox{valign=c}{\includegraphics[width=.1\textwidth]{images/dataset/building-a-winter/thermal.PNG}}  &
                \makecell{84 (train)                                                                                                                                                                                                                              \\ 12 (test)}&
                \makecell{-15.7°C                                                                                                                                                                                                                                 \\ 15.6°C }\\

                Dorm 1                                                                                                    & \adjustbox{valign=c}{\includegraphics[width=.1\textwidth]{images/dataset/LES-dorm1/rgb.png}}    &
                \adjustbox{valign=c}{\includegraphics[width=.1\textwidth]{images/dataset/LES-dorm1/thermal.png}}          &
                \makecell{100 (train)                                                                                                                                                                                                                             \\ 50 (test)}&
                \makecell{-22.1°C*                                                                                                                                                                                                                                \\ 24.6°C }\\

                Dorm 2                                                                                                    & \adjustbox{valign=c}{\includegraphics[width=.1\textwidth]{images/dataset/LES-dorm2/rgb.png}}    &
                \adjustbox{valign=c}{\includegraphics[width=.1\textwidth]{images/dataset/LES-dorm2/thermal.PNG}}          &
                \makecell{73 (train)                                                                                                                                                                                                                              \\ 10 (test)}&
                \makecell{-8.4°C                                                                                                                                                                                                                                  \\ 11.8°C }\\

                MED building                                                                                              & \adjustbox{valign=c}{\includegraphics[width=.1\textwidth]{images/dataset/MED_building/rgb.png}} &
                \adjustbox{valign=c}{\includegraphics[width=.1\textwidth]{images/dataset/MED_building/thermal.png}}       &
                \makecell{93 (train)                                                                                                                                                                                                                              \\ 46 (test)}&
                \makecell{-25.2°C*                                                                                                                                                                                                                                \\ 37.5°C }\\

                INR building                                                                                              & \adjustbox{valign=c}{\includegraphics[width=.1\textwidth]{images/dataset/INR/rgb.png}}          &
                \adjustbox{valign=c}{\includegraphics[width=.1\textwidth]{images/dataset/INR/thermal.png}}                &
                \makecell{111 (train)                                                                                                                                                                                                                             \\ 40 (test)}&
                \makecell{-35.5°C*                                                                                                                                                                                                                                \\ 45.6°C }\\

                BI building                                                                                               & \adjustbox{valign=c}{\includegraphics[width=.1\textwidth]{images/dataset/BI/rgb.png}}           &
                \adjustbox{valign=c}{\includegraphics[width=.1\textwidth]{images/dataset/BI/thermal.png}}                 &
                \makecell{131 (train)                                                                                                                                                                                                                             \\ 50 (test)}&
                \makecell{-0.03°C                                                                                                                                                                                                                                 \\ 19.2°C }\\

                \bottomrule
            \end{tabular}
        }
        \label{tbl:thermoscenes-summary:outdoor}

         &
        \resizebox{.445\textwidth}{!}{%
            \begin{tabular}{l cccc}
                \toprule
                Scene                                                                                                & RGB                                                                                       & Thermal & \#views & \makecell{Temp. \\ range }\\
                \midrule

                \makecell[l]{Hot Water                                                                                                                                                                                                                 \\ Kettle } & \adjustbox{valign=c}{\includegraphics[width=.1\textwidth]{images/dataset/heated-kettle/rgb.JPG}} &
                \adjustbox{valign=c}{\includegraphics[width=.1\textwidth]{images/dataset/heated-kettle/thermal.JPG}} &
                \makecell{77 (train)                                                                                                                                                                                                                   \\ 10 (test)}&
                \makecell{20.0°C                                                                                                                                                                                                                       \\ 87.3°C }\\

                \makecell[l]{Melting                                                                                                                                                                                                                   \\ Ice Cup } & \adjustbox{valign=c}{\includegraphics[width=.1\textwidth]{images/dataset/melting-cup/rgb.JPG}} &
                \adjustbox{valign=c}{\includegraphics[width=.1\textwidth]{images/dataset/melting-cup/thermal.PNG}}   &
                \makecell{85 (train)                                                                                                                                                                                                                   \\ 12 (test)}&
                \makecell{0.4°C                                                                                                                                                                                                                        \\ 25.5°C }\\

                \makecell[l]{ Frozen                                                                                                                                                                                                                   \\ Ice Cup } & \adjustbox{valign=c}{\includegraphics[width=.1\textwidth]{images/dataset/freezing-cup/rgb.JPG}} &
                \adjustbox{valign=c}{\includegraphics[width=.1\textwidth]{images/dataset/freezing-cup/thermal.JPG}}  &
                \makecell{133 (train)                                                                                                                                                                                                                  \\ 19 (test)}&
                \makecell{-16.2°C                                                                                                                                                                                                                      \\ 23.1°C }\\

                \makecell[l]{Raspberry                                                                                                                                                                                                                 \\ Pi } & \adjustbox{valign=c}{\includegraphics[width=.1\textwidth]{images/dataset/raspberrypi/rgb.JPG}} &
                \adjustbox{valign=c}{\includegraphics[width=.1\textwidth]{images/dataset/raspberrypi/thermal.PNG}}   &
                \makecell{111 (train)                                                                                                                                                                                                                  \\ 15 (test)}&
                \makecell{22.3°C                                                                                                                                                                                                                       \\ 41.8°C }\\

                \makecell[l]{Double                                                                                                                                                                                                                    \\ Robot } & \adjustbox{valign=c}{\includegraphics[width=.1\textwidth]{images/dataset/double/rgb.JPG}} &
                \adjustbox{valign=c}{\includegraphics[width=.1\textwidth]{images/dataset/double/thermal.PNG}}        &
                \makecell{83 (train)                                                                                                                                                                                                                   \\ 11 (test)}&
                \makecell{21.0°C                                                                                                                                                                                                                       \\ 29.3°C }\\

                \makecell[l]{Heated                                                                                                                                                                                                                    \\ Water Cup } & \adjustbox{valign=c}{\includegraphics[width=.1\textwidth]{images/dataset/heated-cup/rgb.JPG}} &
                \adjustbox{valign=c}{\includegraphics[width=.1\textwidth]{images/dataset/heated-cup/thermal.JPG}}    &
                \makecell{95 (train)                                                                                                                                                                                                                   \\ 13 (test)}&
                \makecell{23.8°C                                                                                                                                                                                                                       \\ 68.6°C }\\

                Trees                                                                                                & \adjustbox{valign=c}{\includegraphics[width=.1\textwidth]{images/dataset/trees/rgb.JPG}}  &
                \adjustbox{valign=c}{\includegraphics[width=.1\textwidth]{images/dataset/trees/thermal.PNG}}         &
                \makecell{73 (train)                                                                                                                                                                                                                   \\ 10 (test)}&
                \makecell{-8.4°C                                                                                                                                                                                                                       \\ 11.8°C }\\

                \makecell[l]{Laptop }                                                                                & \adjustbox{valign=c}{\includegraphics[width=.1\textwidth]{images/dataset/laptop/rgb.png}} &
                \adjustbox{valign=c}{\includegraphics[width=.1\textwidth]{images/dataset/laptop/thermal.png}}        &
                \makecell{97 (train)                                                                                                                                                                                                                   \\ 50 (test)}&
                \makecell{19.18°C                                                                                                                                                                                                                      \\ 47.92°C }\\


                \bottomrule
            \end{tabular}
        }
        \label{tbl:thermoscenes-summary}
    \end{tabular}
    \footnotesize *Due to the variations of atmospheric radiation in the sky, the readings of thermal cameras are subject to systematic inaccuracies~\cite{KRUCZEK2023128466} and the measurements are outside the camera's operating range.
    We make available the raw data but, in our experiments, the temperature values under the minimum of the operating range of the camera (-20°C) are clipped to -20°C.

\end{table}

%% file: evaluation.tex
\input{results/outdoor}

\section{Experiments}
\label{sec:evaluation}

\input{results/indoor}

We conduct a comprehensive evaluation of rendered novel views using ThermoScenes.
This section details the evaluation metrics, compares baselines, and discusses results.

\subsection{Evaluation Metrics}

\noindent\textbf{Temperature Metrics.} We assess the accuracy of the rendered temperatures by calculating the Mean Absolute Error (MAE).
Given that thermal images often display uniform temperature distributions outside the regions of interest (ROI), calculating the MAE across the entire image may disproportionately emphasize the ambient background temperature.
From an application standpoint, the temperature of the ROI (e.g., the building facade) is crucial for temperature assessment~\cite{rs10091401}.
Therefore, we also report the $\text{MAE}_{\text{roi}}$, the MAE computed over the region of interest.
We use Otsu's method \cite{4310076} to determine the optimal threshold that distinguishes the ROI pixels from the background in thermal images. We also report the standard image quality metrics to evaluate the reconstruction quality for RGB and thermal images.

\noindent\textbf{Image Quality Metrics.} We report the commonly used image quality metrics Peak-Signal-to-Noise-Ratio (PSNR)~\cite{5596999} and Structural Similarity Metric (SSIM)~\cite{article_ssim} for both thermal and RGB modalities.
For the RGB views, we also include the Learned Perceptual Image Patch Similarity (LPIPS)~\cite{zhang2018unreasonable}; since the LPIPS is specifically designed to evaluate the human-perceived similarity for RGB images, we do not include it for thermal images evaluation.

Note that most works in RGB+thermal novel view synthesis have focused on evaluating only image quality metrics, either solely for the RGB modality \cite{xu2024leveraging,ye2024thermal}---highlighting the visual improvement in RGB brought by the use of thermal images---or on both RGB and thermal modalities \cite{lin2024thermalnerf,ozer2024exploring}.
In both cases, the visual quality of the generated images is being evaluated.
However, the differences between the estimated and ground-truth temperature values is not measured nor analyzed.
In the present work, we rather focus on temperature metrics---while ensuring good visual quality---as accurately reconstructing the surface temperature is crucial for real-world applications such as building envelope analysis for retrofit purposes.


\subsection{Baselines}

We conduct experiments to evaluate (1) the necessity of multimodality, as thermal images alone cannot accurately estimate the scene density due to lack of texture details, and (2) the decoupling of the RGB and thermal modalities, stemming from the physical independence between temperature and color information. Given that ThermoNeRF is based on Nerfacto, we define two baseline methods derived from Nerfacto: Nerfacto$_{\text{th}}$ and Nerfacto$_{\text{rgb+th}}$.

\noindent\textbf{Nerfacto$_{\text{th}}$} is trained with thermal inputs only and processes thermal images as single-channel grayscale images through Nerfacto's standard pipeline.
It highlights the importance of incorporating the RGB modality.

\noindent\textbf{Nerfacto$_{\text{rgb+th}}$}
takes both RGB and thermal modalities as inputs by concatenating them into four-channel images and optimizes the concatenated RGB-thermal images without employing separate MLPs for optimizing each modality.
Tests against Nerfacto$_{\text{rgb+th}}$ highlights the importance of disjoint optimization of the modalities.

\subsection{Implementation Details}

We use 2-layer MLP with width 64 and 1 output dimension for MLP$_{\text{th}}$ and MLP$_{\text{dens}}$. For MLP$_{\text{rgb}}$, we use 3-layer MLP with width 64 and 3 output dimensions.
We use ReLU activation function for all MLPs and Sigmoid function for the head of MLP$_{\text{th}}$ and MLP$_{\text{rgb}}$.
We use the default NeRFstudio \cite{nerfstudio} settings for the hash encoder which has a minimum and maximum resolution of 16 and 1024 respectively and 2 features per level.
Additionally, we use the default settings of the spherical harmonic encoding with 4 spherical harmonic levels.
All details are available in our code repository
\footnote{\url{https://github.com/Schindler-EPFL-Lab/thermo-nerf}}.

We train ThermoNeRF as well as our baselines for 30k iterations with 4096 rays per batch. We use a learning rate of $10^{-2}$ with exponential decay decreasing the learning rate by an order of magnitude to $10^{-3}$.
The total training time is around 40 minutes using a single T4 GPU.

\subsection{Thermal View Synthesis}

\begin{figure}[t]
  \centering
  \begin{subfigure}[t]{0.24\textwidth}
    \centering
    \includegraphics[width=.95\textwidth]{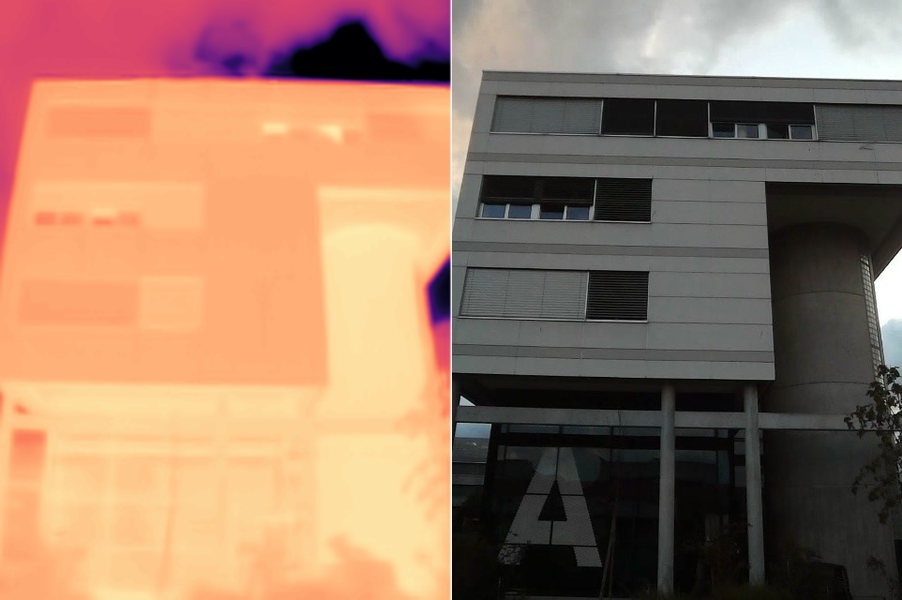}
  \end{subfigure}%
  \begin{subfigure}[t]{0.15\textwidth}
    \centering
    \includegraphics[width=.75\textwidth]{images/building-spring-coloured/nerfacto_t.png}
  \end{subfigure}%
  \begin{subfigure}[t]{0.24\textwidth}
    \centering
    \includegraphics[width=.95\textwidth]{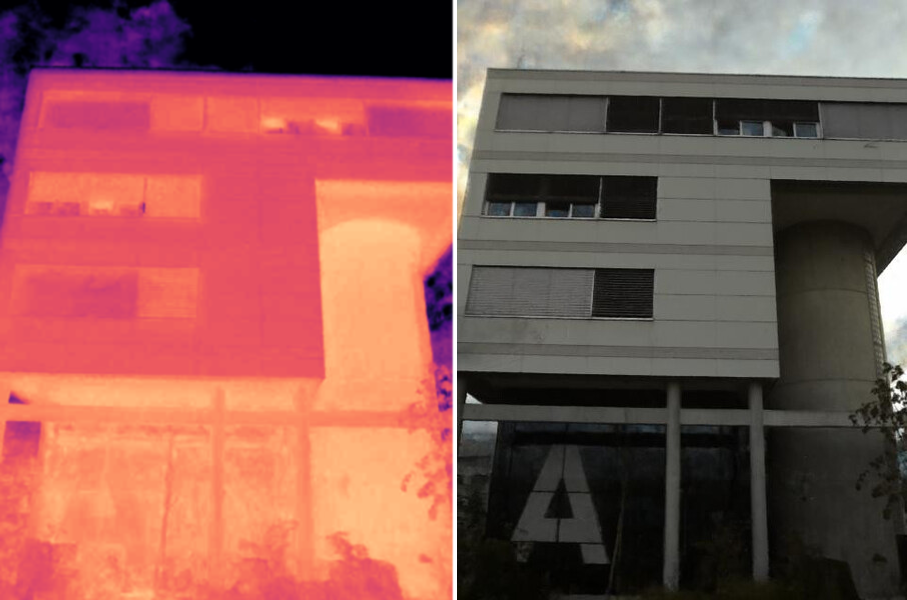}
  \end{subfigure}%
  \begin{subfigure}[t]{0.24\textwidth}
    \centering
    \includegraphics[width=.95\textwidth]{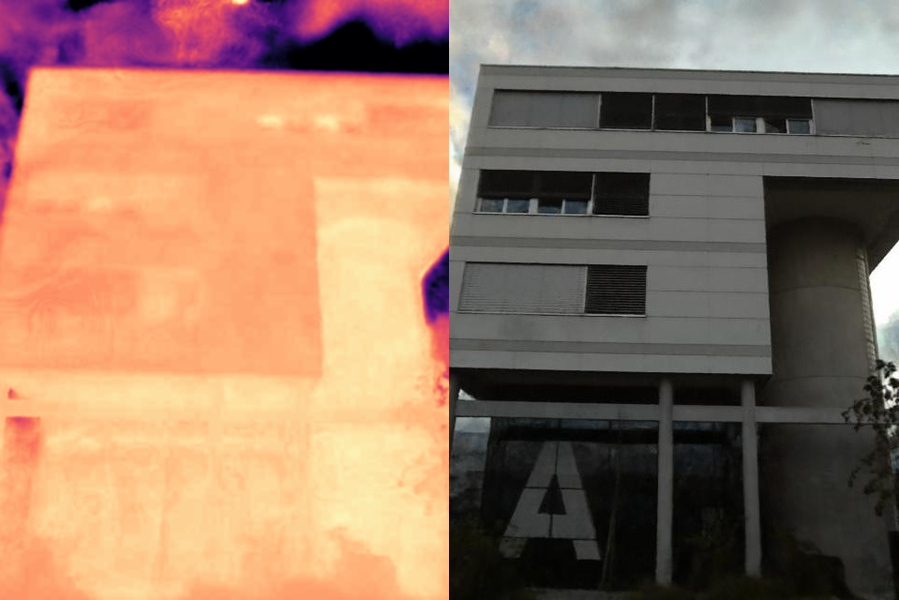}
  \end{subfigure}%
  \begin{subfigure}[t]{0.075\textwidth}
    \centering
    \includegraphics[height=.1\textheight]{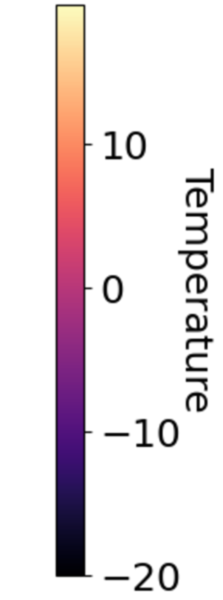}
  \end{subfigure}
  \\
  \begin{subfigure}[t]{0.24\textwidth}
    \centering
    \includegraphics[width=.95\textwidth]{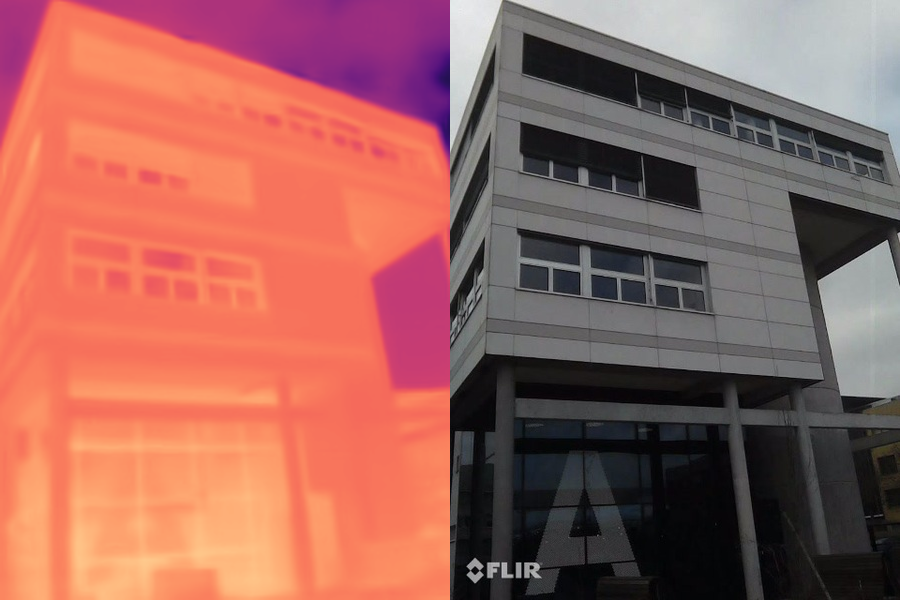}
  \end{subfigure}%
  \begin{subfigure}[t]{0.15\textwidth}
    \centering
    \includegraphics[width=.75\textwidth]{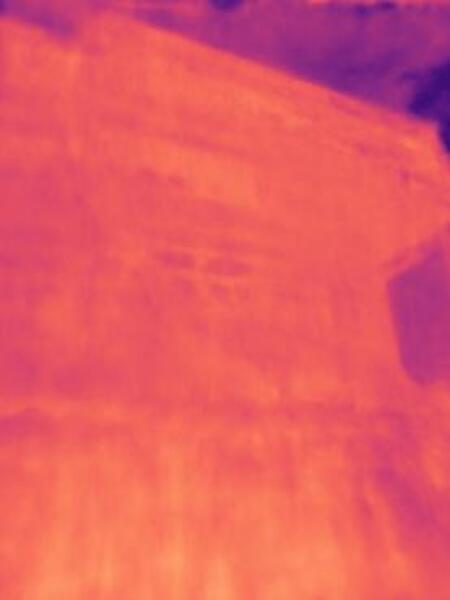}
  \end{subfigure}%
  \begin{subfigure}[t]{0.24\textwidth}
    \centering
    \includegraphics[width=.95\textwidth]{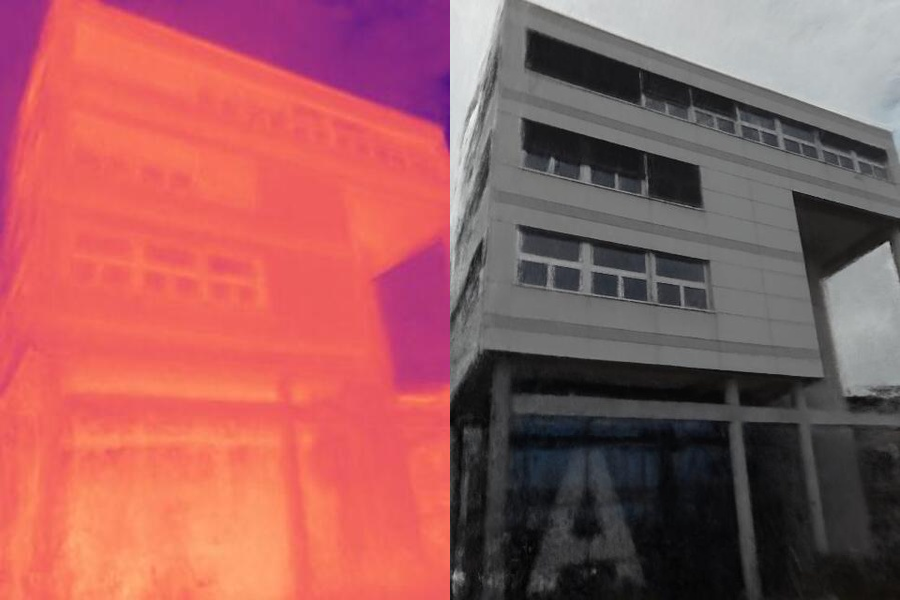}
  \end{subfigure}%
  \begin{subfigure}[t]{0.24\textwidth}
    \centering
    \includegraphics[width=.95\textwidth]{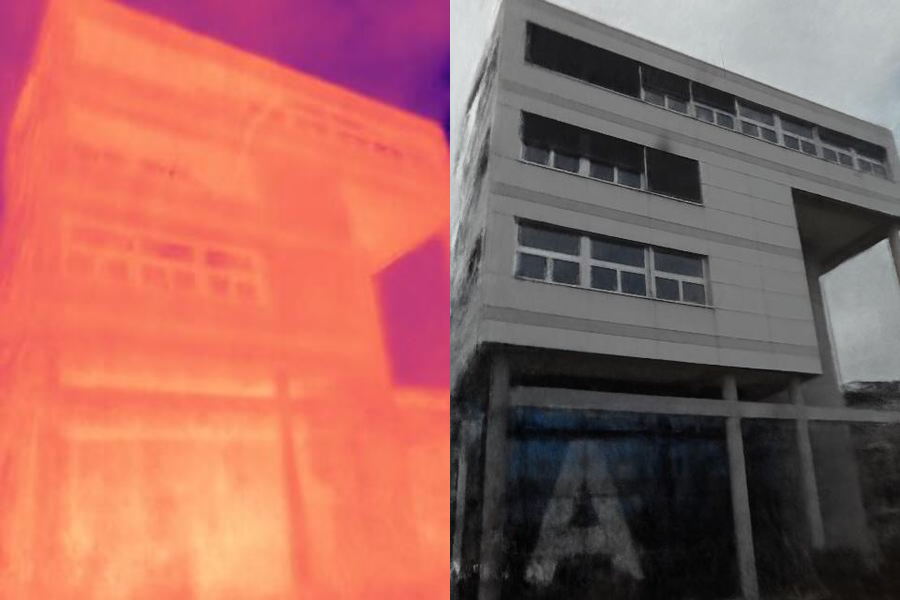}
  \end{subfigure}%
  \begin{subfigure}[t]{0.075\textwidth}
    \centering
    \includegraphics[height=.1\textheight]{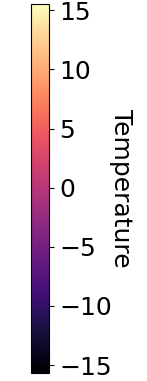}
  \end{subfigure}
  \\
  \begin{subfigure}[t]{0.24\textwidth}
    \centering
    \includegraphics[width=.95\textwidth]{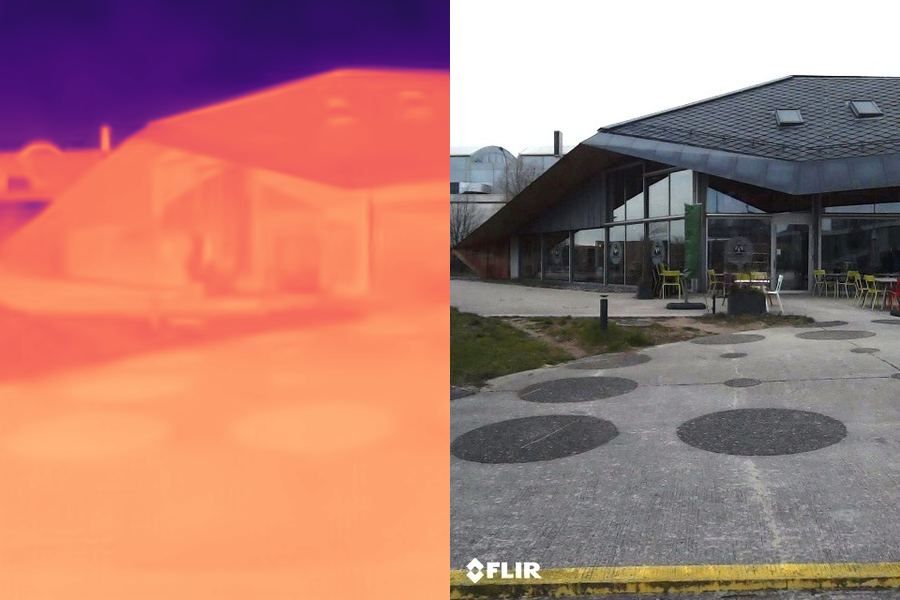}
  \end{subfigure}%
  \begin{subfigure}[t]{0.15\textwidth}
    \centering
    \includegraphics[width=.75\textwidth]{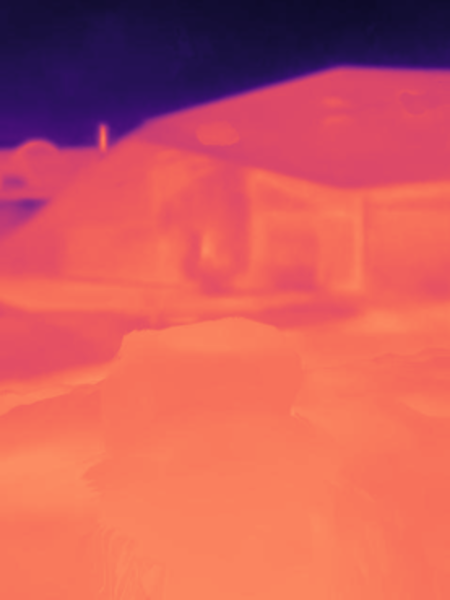}
  \end{subfigure}%
  \begin{subfigure}[t]{0.24\textwidth}
    \centering
    \includegraphics[width=.95\textwidth]{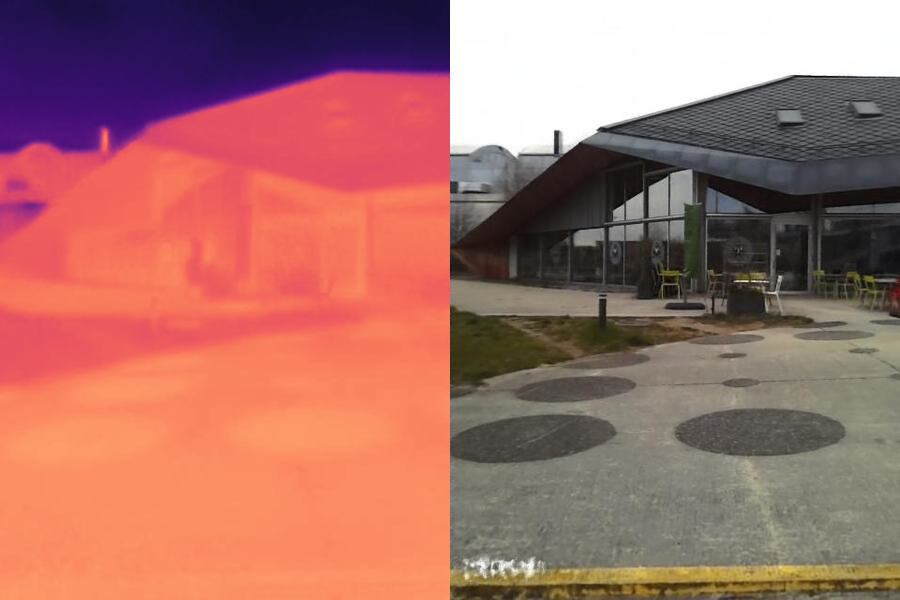}
  \end{subfigure}%
  \begin{subfigure}[t]{0.24\textwidth}
    \centering
    \includegraphics[width=.95\textwidth]{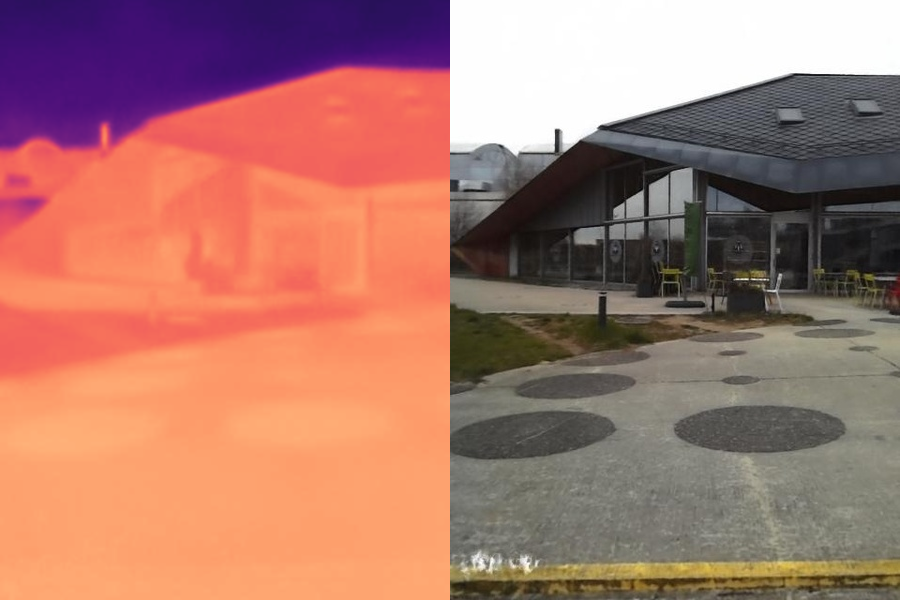}
  \end{subfigure}%
  \begin{subfigure}[t]{0.075\textwidth}
    \centering
    \includegraphics[height=.095\textheight]{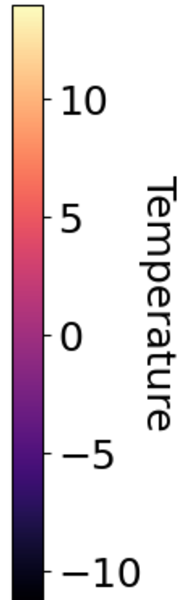}
  \end{subfigure}
  \\
  \begin{subfigure}[t]{0.24\textwidth}
    \centering
    \includegraphics[width=.95\textwidth]{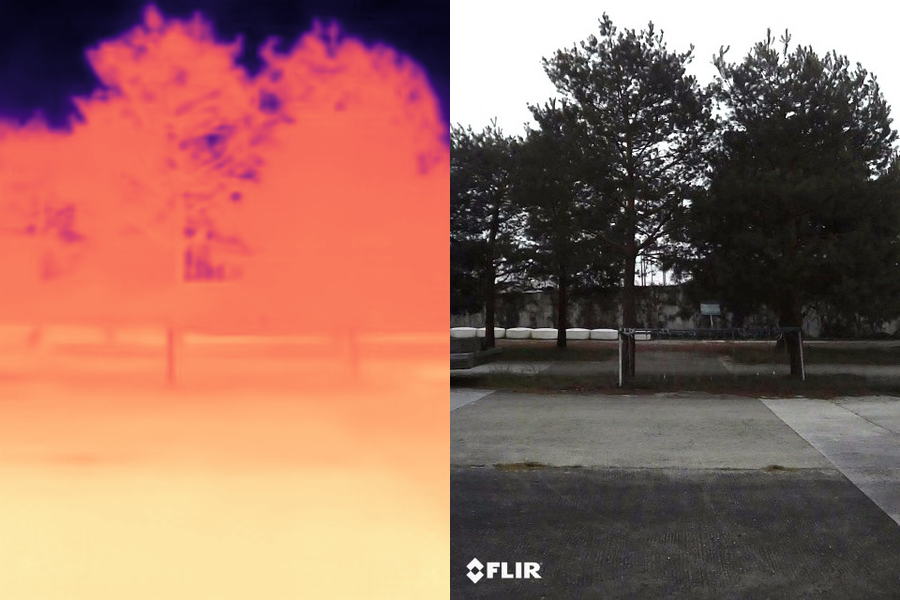}
  \end{subfigure}%
  \begin{subfigure}[t]{0.15\textwidth}
    \centering
    \includegraphics[width=.75\textwidth]{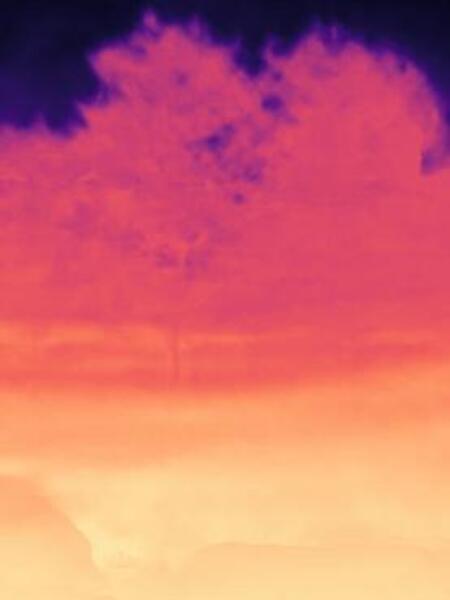}
  \end{subfigure}%
  \begin{subfigure}[t]{0.24\textwidth}
    \centering
    \includegraphics[width=.95\textwidth]{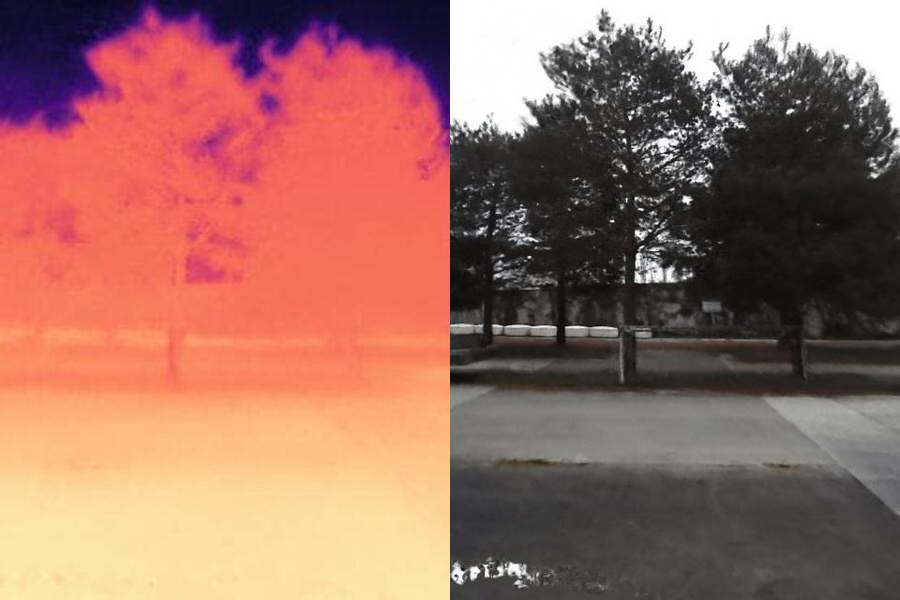}
  \end{subfigure}%
  \begin{subfigure}[t]{0.24\textwidth}
    \centering
    \includegraphics[width=.95\textwidth]{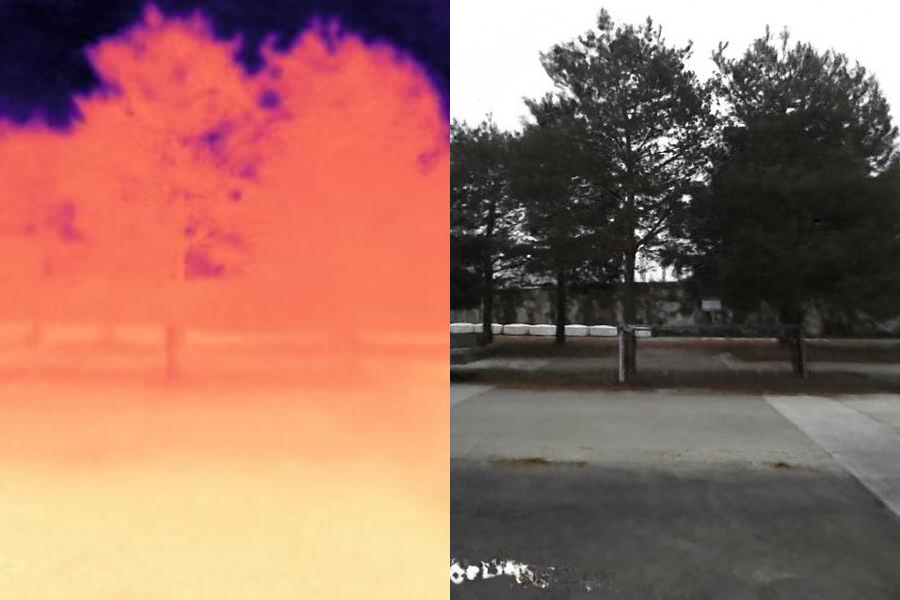}
  \end{subfigure}%
  \begin{subfigure}[t]{0.075\textwidth}
    \centering
    \includegraphics[height=.1\textheight]{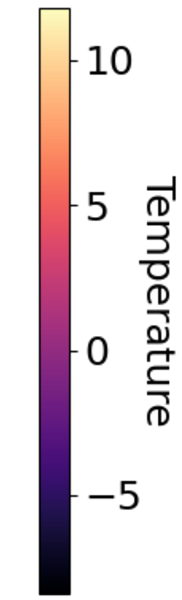}
  \end{subfigure}
  \\

  \begin{subfigure}[t]{0.24\textwidth}
    \centering
    \includegraphics[width=.95\textwidth]{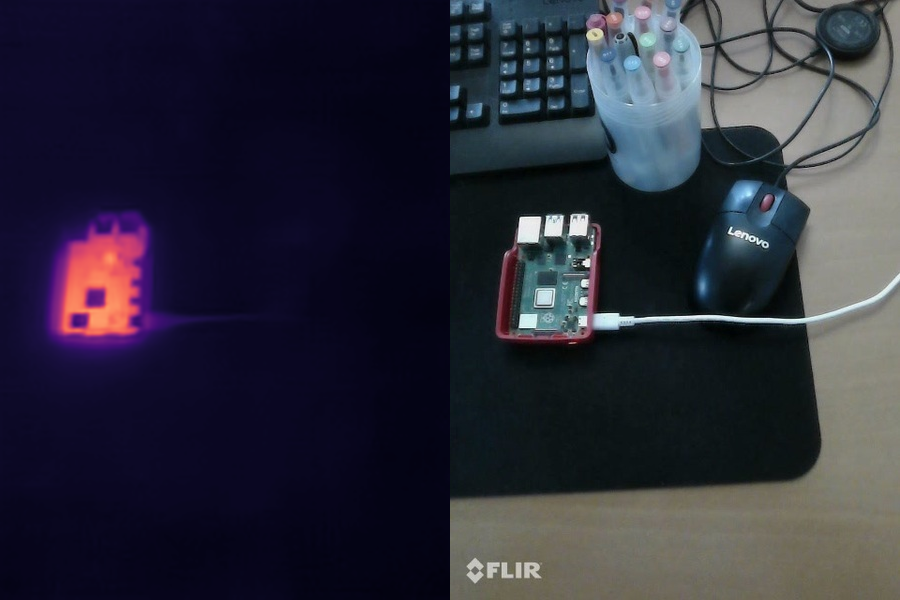}
    \caption{Ground truth}
  \end{subfigure}%
  \begin{subfigure}[t]{0.15\textwidth}
    \centering
    \includegraphics[width=.75\textwidth]{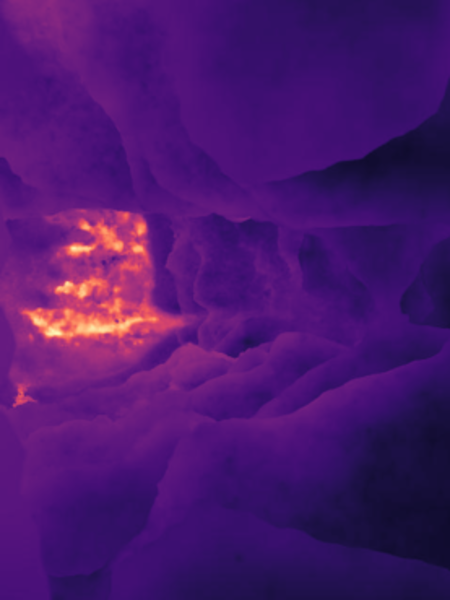}
    \caption{Nerfacto$_\text{th}$}
    \label{fig:thermal-only}
  \end{subfigure}%
  \begin{subfigure}[t]{0.24\textwidth}
    \centering
    \includegraphics[width=.95\textwidth]{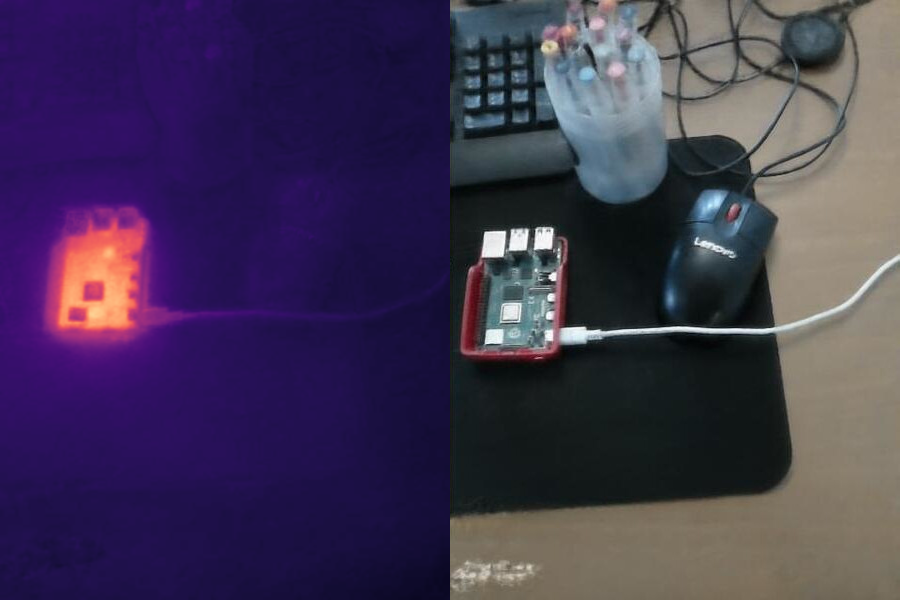}
    \caption{Nerfacto$_\text{rgb+th}$}
    \label{subfig:n_rgb_t_rec_supp}
  \end{subfigure}%
  \begin{subfigure}[t]{0.24\textwidth}
    \centering
    \includegraphics[width=.95\textwidth]{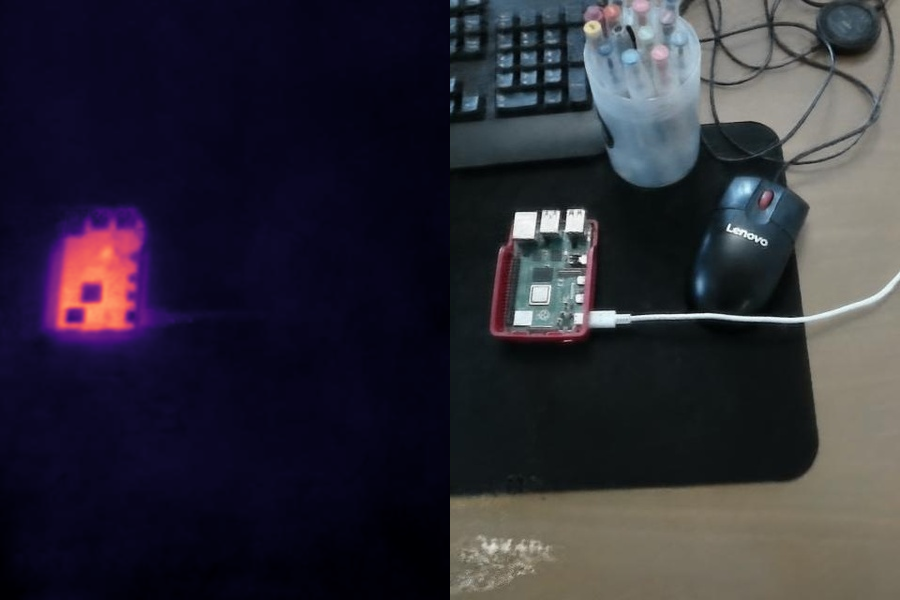}
    \caption{ThermoNeRF}
  \end{subfigure}%
  \begin{subfigure}[t]{0.075\textwidth}
    \centering
    \includegraphics[height=.1\textheight]{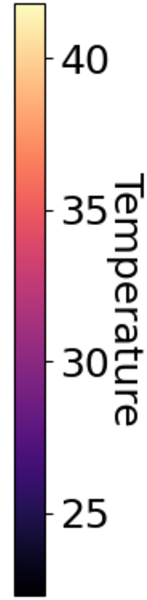}
  \end{subfigure}
  \\
  \caption{Comparison of examples of thermal and RGB renderings of unseen poses for the four outdoor scenes (top to bottom rows): Building (Spring), Building (Winter), Exhibition Building, Trees, and Raspberry pi.
    ThermoNeRF is closest to the ground-truth thermal image, while preserving RGB quality.}
  \label{fig:results-comparison-four}
\end{figure}

\cref{tab:thermal-results-indoor} and \cref{tab:thermal-results-outdoor} provide a comparative analysis of our method's performance for novel view synthesis for the thermal modality on the ThermoScenes dataset.

\begin{figure}[t]
  \centering
  \scriptsize
  \begin{subfigure}[t]{0.15\textwidth}
    \centering
    \includegraphics[width=.7\textwidth]{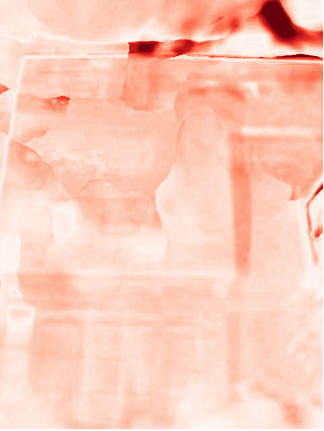}
    MAE=4.67°C
  \end{subfigure}%
  \begin{subfigure}[t]{0.15\textwidth}
    \centering
    \includegraphics[width=.7\textwidth]{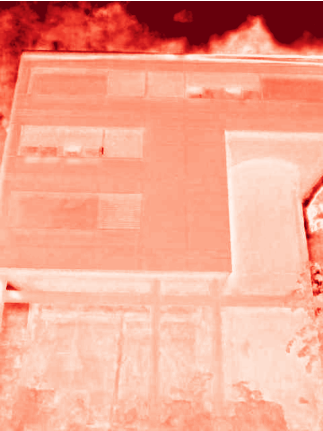}
    MAE=8.32°C
  \end{subfigure}%
  \begin{subfigure}[t]{0.15\textwidth}
    \centering
    \includegraphics[width=.7\textwidth]{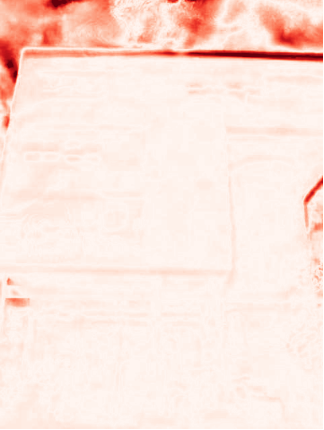}
    MAE=1.71°C
  \end{subfigure}%
  \begin{subfigure}[t]{0.05\textwidth}
    \centering
    \includegraphics[height=.1\textheight]{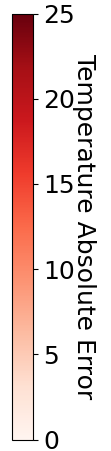}
  \end{subfigure}
  \begin{subfigure}[t]{0.15\textwidth}
    \centering
    \includegraphics[width=.7\textwidth]{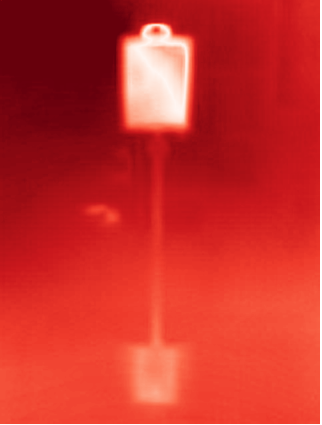}
    MAE=7.71°C
  \end{subfigure}%
  \begin{subfigure}[t]{0.15\textwidth}
    \centering
    \includegraphics[width=.7\textwidth]{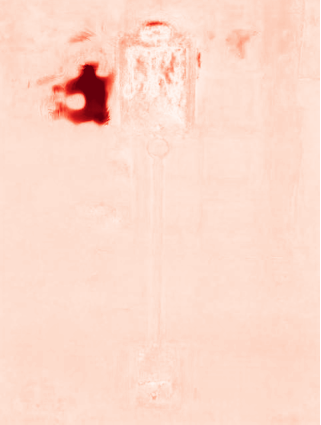}
    MAE=1.48°C
  \end{subfigure}%
  \begin{subfigure}[t]{0.15\textwidth}
    \centering
    \includegraphics[width=.7\textwidth]{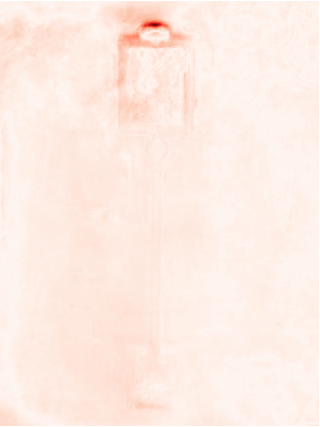}
    MAE=0.50°C
  \end{subfigure}%
  \begin{subfigure}[t]{0.05\textwidth}
    \centering
    \includegraphics[height=.1\textheight]{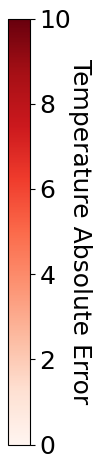}
  \end{subfigure}%
  \\
  \begin{subfigure}[t]{0.15\textwidth}
    \centering
    \includegraphics[width=.7\textwidth]{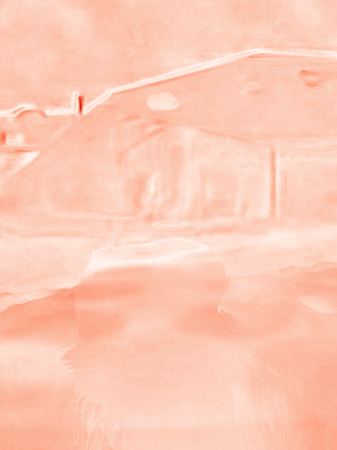}
    MAE=1.97°C
  \end{subfigure}%
  \begin{subfigure}[t]{0.15\textwidth}
    \centering
    \includegraphics[width=.7\textwidth]{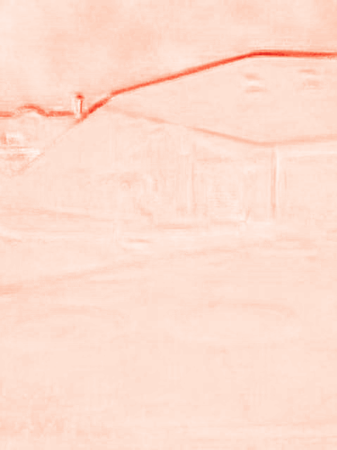}
    MAE=1.33°C
  \end{subfigure}%
  \begin{subfigure}[t]{0.15\textwidth}
    \centering
    \includegraphics[width=.7\textwidth]{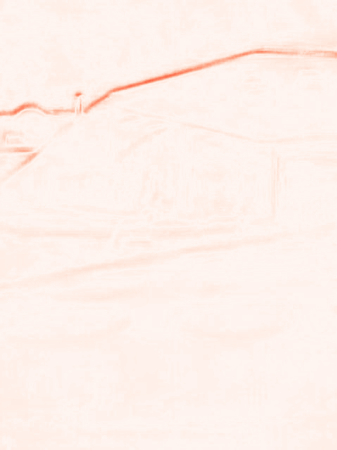}
    MAE=0.21°C
  \end{subfigure}%
  \begin{subfigure}[t]{0.05\textwidth}
    \centering
    \includegraphics[height=.1\textheight]{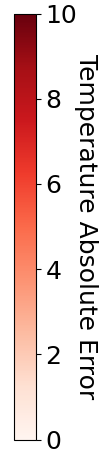}
  \end{subfigure}%
  \begin{subfigure}[t]{0.15\textwidth}
    \centering
    \includegraphics[width=.7\textwidth]{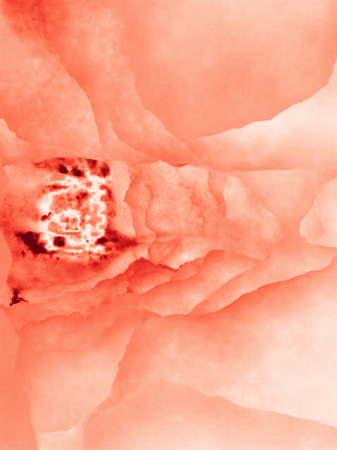}
    MAE=3.27°C
  \end{subfigure}%
  \begin{subfigure}[t]{0.15\textwidth}
    \centering
    \includegraphics[width=.7\textwidth]{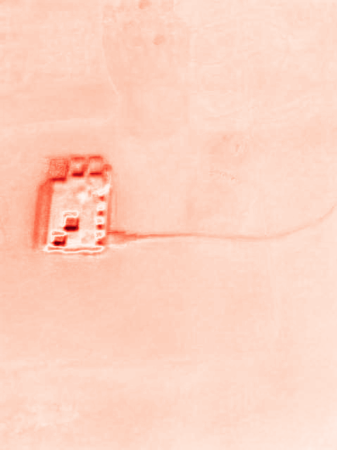}
    MAE=1.68°C
  \end{subfigure}%
  \begin{subfigure}[t]{0.15\textwidth}
    \centering
    \includegraphics[width=.7\textwidth]{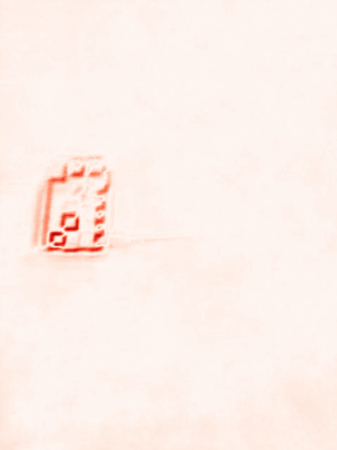}
    MAE=0.25°C
  \end{subfigure}%
  \begin{subfigure}[t]{0.05\textwidth}
    \centering
    \includegraphics[height=.1\textheight]{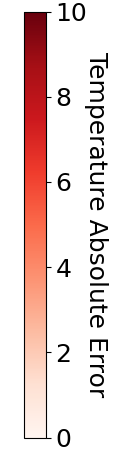}
  \end{subfigure}%
  \\
  \begin{subfigure}[t]{0.15\textwidth}
    \centering
    \includegraphics[width=.7\textwidth]{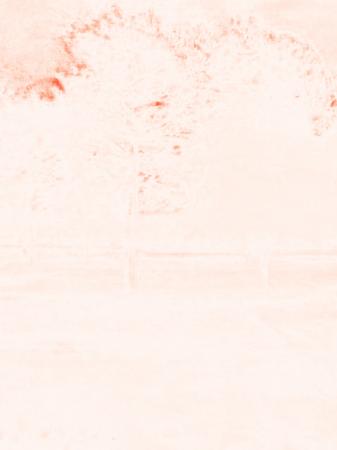}
    MAE=0.32°C
    \caption{N$_\text{th}$}
  \end{subfigure}%
  \begin{subfigure}[t]{0.15\textwidth}
    \centering
    \includegraphics[width=.7\textwidth]{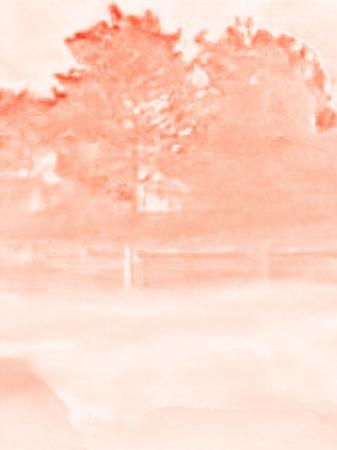}
    MAE=1.50°C
    \caption{N$_\text{rgb+th}$}
  \end{subfigure}%
  \begin{subfigure}[t]{0.15\textwidth}
    \centering
    \includegraphics[width=.7\textwidth]{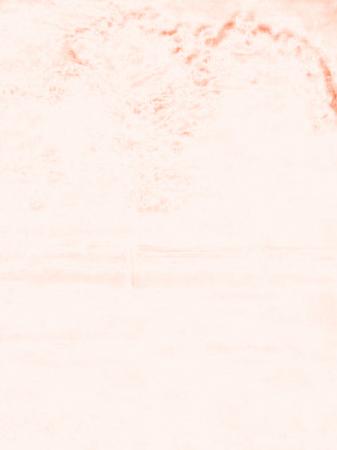}
    MAE=0.23°C
    \caption{\tiny ThermoNeRF}
  \end{subfigure}%
  \begin{subfigure}[t]{0.05\textwidth}
    \centering
    \includegraphics[height=.1\textheight]{images/raspberrypi_coloured/mae_cmap.png}
  \end{subfigure}%
  \begin{subfigure}[t]{0.15\textwidth}
    \centering
    \includegraphics[width=.7\textwidth]{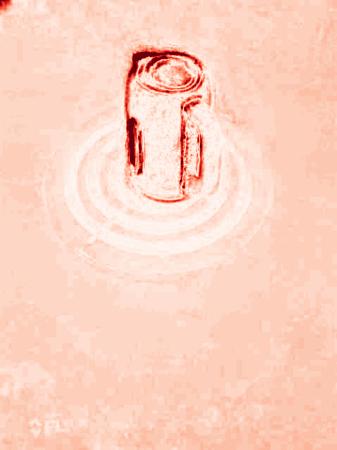}
    MAE=1.69°C
    \caption{N$_\text{th}$}
  \end{subfigure}%
  \begin{subfigure}[t]{0.15\textwidth}
    \centering
    \includegraphics[width=.7\textwidth]{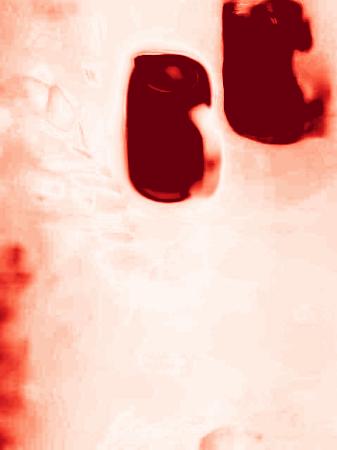}
    MAE=3.50°C
    \caption{N$_\text{rgb+th}$}
  \end{subfigure}%
  \begin{subfigure}[t]{0.15\textwidth}
    \centering
    \includegraphics[width=.7\textwidth]{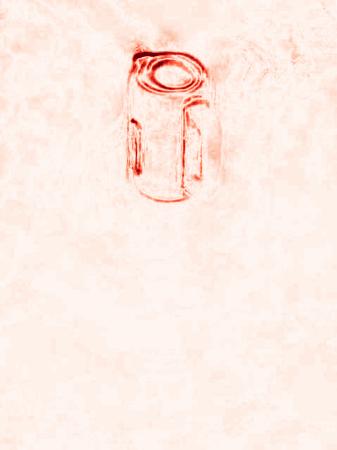}
    MAE=0.43°C
    \caption{\tiny ThermoNeRF}
  \end{subfigure}%
  \begin{subfigure}[t]{0.05\textwidth}
    \centering
    \includegraphics[height=.1\textheight]{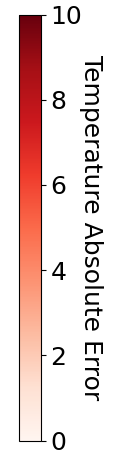}
  \end{subfigure}%
  \caption{Per-pixel absolute errors in temperature estimation for renderings of unseen poses for three outdoor scenes: (left top to bottom) Building Spring, Exhibition Building and Trees, and three indoor scenes: (right top to bottom) Double Robot, RaspberryPi, and Heated Water Kettle.
    We observe fewer errors on ThermoNeRF than the baselines. Note that N$_\text{rgb+th}$ stands for Nerfacto$_\text{rgb+th}$.
  }
  \label{fig:errors-comparison}
  \vspace{-3mm}
\end{figure}

With an average MAE of $1.13^\circ$C on building scenes and $0.41^\circ$C on other scenes, ThermoNeRF significantly outperforms $\text{Nerfacto}_{\text{rgb+th}}$ ($2.17^\circ$C and $0.91^\circ$C) and $\text{Nerfacto}_{\text{th}}$ ($3.11^\circ$C and $2.41^\circ$C).
When trained on thermal images alone, $\text{Nerfacto}_{\text{th}}$ fails to learn an accurate representation of the scene (as visible in \cref{fig:thermal-only}) and has the lowest image quality and temperature estimation results.
On the other hand, thanks to the concatenation of RGB and thermal information, $\text{Nerfacto}_{\text{rgb+th}}$ obtains better results than $\text{Nerfacto}_{\text{th}}$: MAE$_{\text{roi}}$ of $1.79^\circ$C, against  $3.09^\circ$C for $\text{Nerfacto}_{\text{th}}$ on building scenes, and $2.96^\circ$C against $7.30^\circ$C on other scenes.
However, the estimated temperatures are impacted by the joint optimization of both modalities in the same MLP, as evidenced in \cref{fig:errors-comparison} where the error in temperature prediction is high on the ROI.
On the contrary, the average MAE$_{\text{roi}}$ and MAE for ThermoNerf mark a significant improvement over the other methods (for the MAE $1.13^\circ$C on building scenes and $1.57^\circ$C for others and for the MAE$_{\text{roi}}$ $1.04^\circ$C for buildings and $1.57^\circ$C for others).



\begin{figure}[t]
  \centering
  \scriptsize
  \begin{subfigure}[t]{0.15\textwidth}
    \centering
    \includegraphics[width=.7\textwidth]{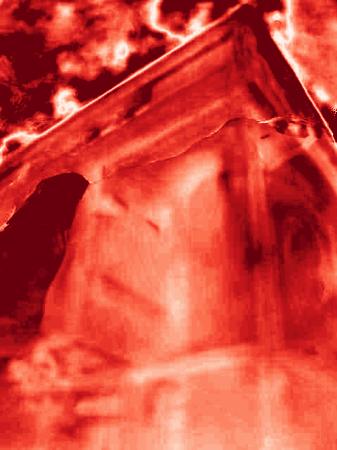}
    MAE=6.60°C
  \end{subfigure}%
  \begin{subfigure}[t]{0.15\textwidth}
    \centering
    \includegraphics[width=.7\textwidth]{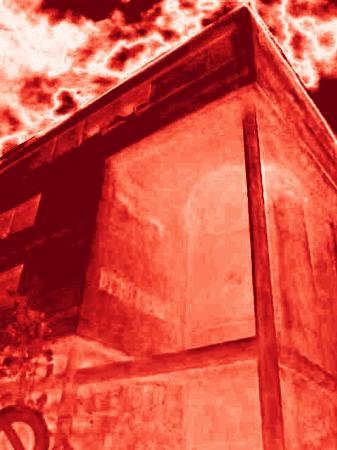}
    MAE=6.26°C
  \end{subfigure}%
  \begin{subfigure}[t]{0.15\textwidth}
    \centering
    \includegraphics[width=.7\textwidth]{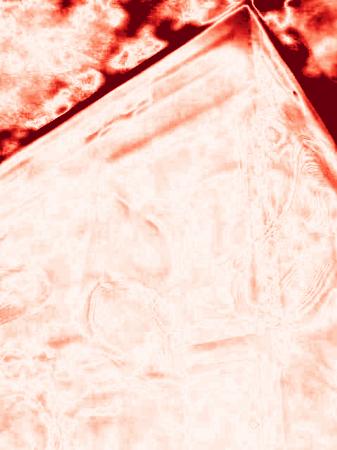}
    MAE=1.83°C
  \end{subfigure}%
  \begin{subfigure}[t]{0.05\textwidth}
    \centering
    \includegraphics[height=.1\textheight]{images/mae_results/heated_water_kettle/freezing_cup_cmap.png}
  \end{subfigure}
  \begin{subfigure}[t]{0.15\textwidth}
    \centering
    \includegraphics[width=.7\textwidth]{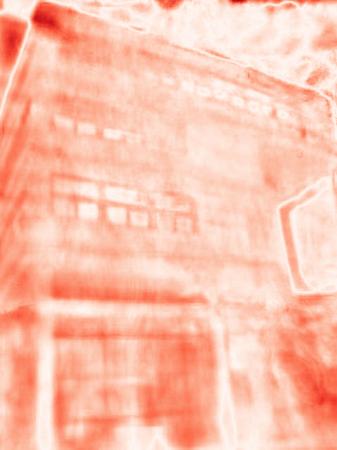}
    MAE=2.20°C
  \end{subfigure}%
  \begin{subfigure}[t]{0.15\textwidth}
    \centering
    \includegraphics[width=.7\textwidth]{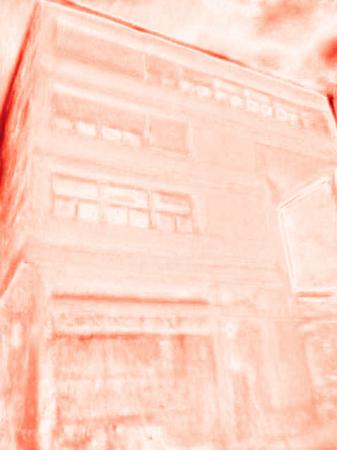}
    MAE=1.57°C
  \end{subfigure}%
  \begin{subfigure}[t]{0.15\textwidth}
    \centering
    \includegraphics[width=.7\textwidth]{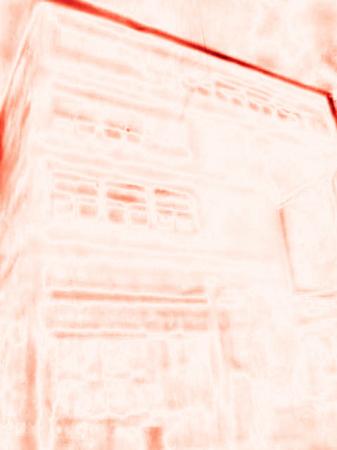}
    MAE=0.72°C
  \end{subfigure}%
  \begin{subfigure}[t]{0.05\textwidth}
    \centering
    \includegraphics[height=.1\textheight]{images/mae_results/heated_water_kettle/freezing_cup_cmap.png}

  \end{subfigure}%
  \\
  \begin{subfigure}[t]{0.15\textwidth}
    \centering
    \includegraphics[width=.7\textwidth]{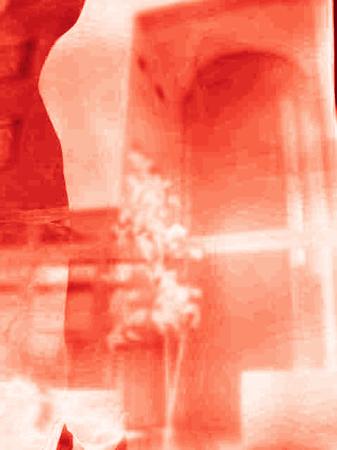}
    MAE=3.81°C
  \end{subfigure}%
  \begin{subfigure}[t]{0.15\textwidth}
    \centering
    \includegraphics[width=.7\textwidth]{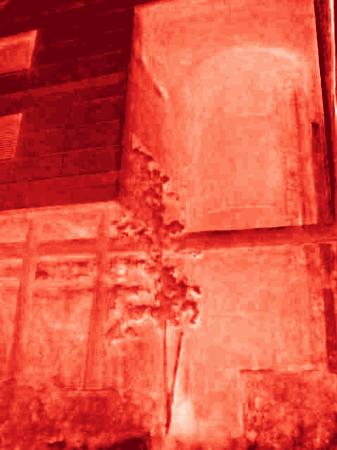}
    MAE=5.91°C
  \end{subfigure}%
  \begin{subfigure}[t]{0.15\textwidth}
    \centering
    \includegraphics[width=.7\textwidth]{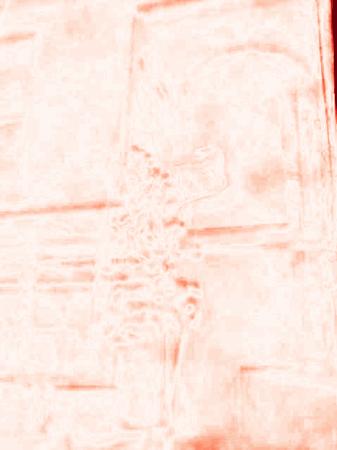}
    MAE=0.68°C
  \end{subfigure}%
  \begin{subfigure}[t]{0.05\textwidth}
    \centering
    \includegraphics[height=.1\textheight]{images/mae_results/heated_water_kettle/freezing_cup_cmap.png}
  \end{subfigure}
  \begin{subfigure}[t]{0.15\textwidth}
    \centering
    \includegraphics[width=.7\textwidth]{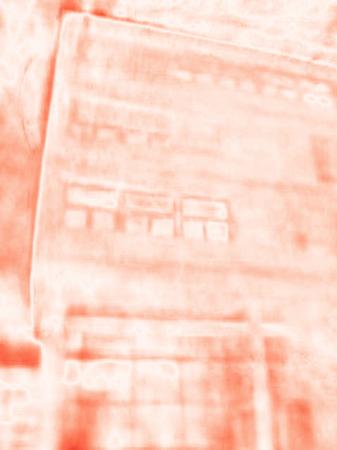}
    MAE=1.45°C
  \end{subfigure}%
  \begin{subfigure}[t]{0.15\textwidth}
    \centering
    \includegraphics[width=.7\textwidth]{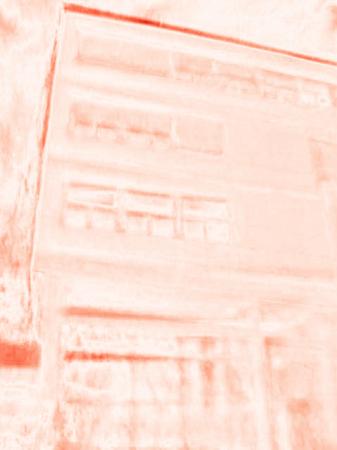}
    MAE=1.03°C
  \end{subfigure}%
  \begin{subfigure}[t]{0.15\textwidth}
    \centering
    \includegraphics[width=.7\textwidth]{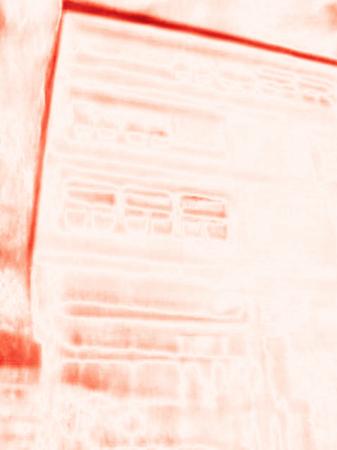}
    MAE=0.81°C
  \end{subfigure}%
  \begin{subfigure}[t]{0.05\textwidth}
    \centering
    \includegraphics[height=.1\textheight]{images/mae_results/heated_water_kettle/freezing_cup_cmap.png}
  \end{subfigure}%
  \\
  \begin{subfigure}[t]{0.15\textwidth}
    \centering
    \includegraphics[width=.7\textwidth]{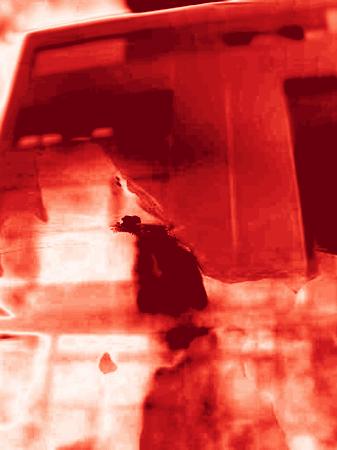}
    MAE=6.55°C
  \end{subfigure}%
  \begin{subfigure}[t]{0.15\textwidth}
    \centering
    \includegraphics[width=.7\textwidth]{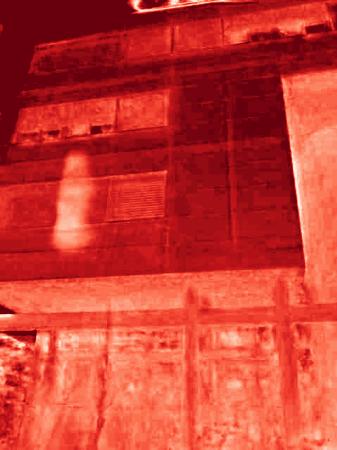}
    MAE=7.25°C
  \end{subfigure}%
  \begin{subfigure}[t]{0.15\textwidth}
    \centering
    \includegraphics[width=.7\textwidth]{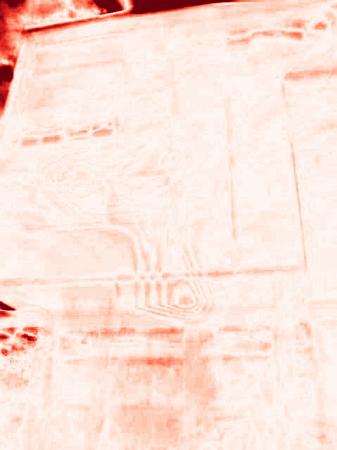}
    MAE=0.96°C
  \end{subfigure}%
  \begin{subfigure}[t]{0.05\textwidth}
    \centering
    \includegraphics[height=.1\textheight]{images/mae_results/heated_water_kettle/freezing_cup_cmap.png}
  \end{subfigure}
  \begin{subfigure}[t]{0.15\textwidth}
    \centering
    \includegraphics[width=.7\textwidth]{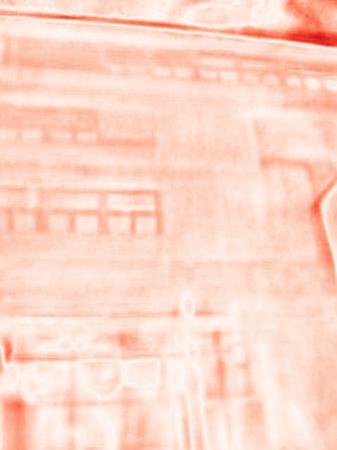}
    MAE=1.47°C
  \end{subfigure}%
  \begin{subfigure}[t]{0.15\textwidth}
    \centering
    \includegraphics[width=.7\textwidth]{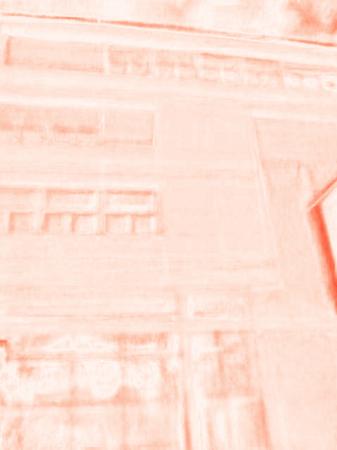}
    MAE=1.10°C
  \end{subfigure}%
  \begin{subfigure}[t]{0.15\textwidth}
    \centering
    \includegraphics[width=.7\textwidth]{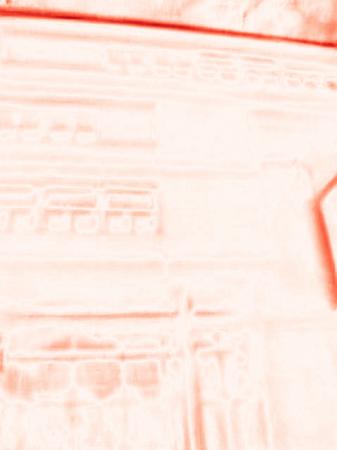}
    MAE=0.62°C
  \end{subfigure}%
  \begin{subfigure}[t]{0.05\textwidth}
    \centering
    \includegraphics[height=.1\textheight]{images/mae_results/heated_water_kettle/freezing_cup_cmap.png}
  \end{subfigure}%
  \\
  \begin{subfigure}[t]{0.15\textwidth}
    \centering
    \includegraphics[width=.7\textwidth]{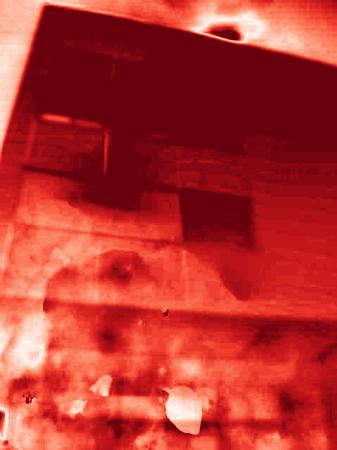}
    MAE=7.68°C
    \caption{N$_\text{th}$}
  \end{subfigure}%
  \begin{subfigure}[t]{0.15\textwidth}
    \centering
    \includegraphics[width=.7\textwidth]{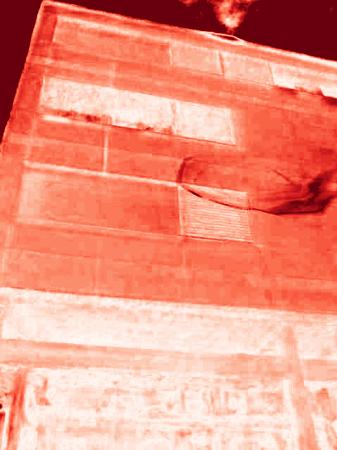}
    MAE=4.46°C
    \caption{N$_\text{rgb+th}$}
  \end{subfigure}%
  \begin{subfigure}[t]{0.15\textwidth}
    \centering
    \includegraphics[width=.7\textwidth]{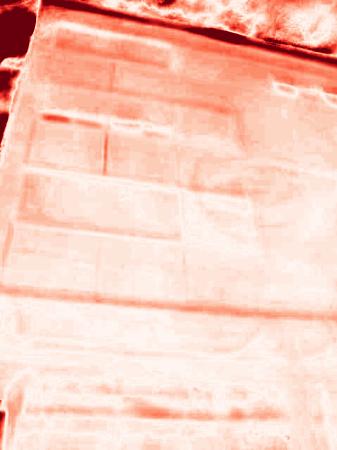}
    MAE=1.84°C
    \caption{\tiny ThermoNeRF}
  \end{subfigure}%
  \begin{subfigure}[t]{0.05\textwidth}
    \centering
    \includegraphics[height=.1\textheight]{images/mae_results/heated_water_kettle/freezing_cup_cmap.png}
  \end{subfigure}%
  \begin{subfigure}[t]{0.15\textwidth}
    \centering
    \includegraphics[width=.7\textwidth]{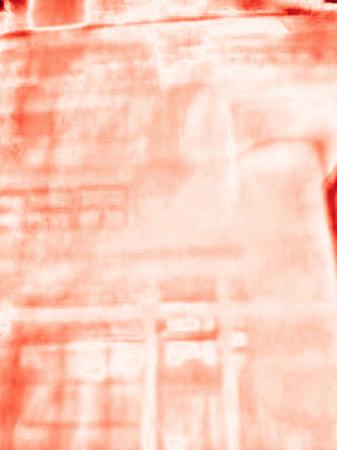}
    MAE=1.80°C
    \caption{N$_\text{th}$}
  \end{subfigure}%
  \begin{subfigure}[t]{0.15\textwidth}
    \centering
    \includegraphics[width=.7\textwidth]{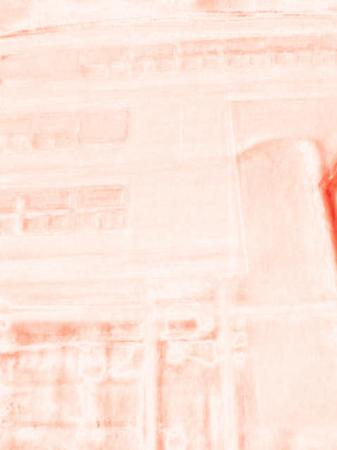}
    MAE=0.83°C
    \caption{N$_\text{rgb+th}$}
  \end{subfigure}%
  \begin{subfigure}[t]{0.15\textwidth}
    \centering
    \includegraphics[width=.7\textwidth]{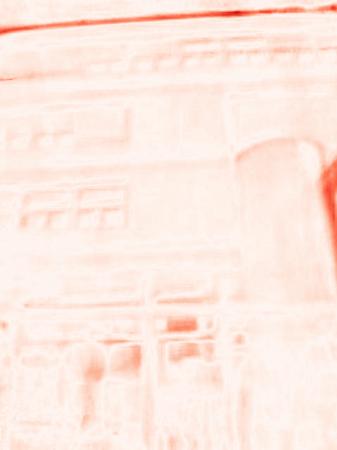}
    MAE=0.60°C
    \caption{\tiny ThermoNeRF}
  \end{subfigure}%
  \begin{subfigure}[t]{0.05\textwidth}
    \centering
    \includegraphics[height=.1\textheight]{images/mae_results/heated_water_kettle/freezing_cup_cmap.png}
  \end{subfigure}%

  \caption{Comparison of per-pixel absolute errors in temperature estimation for renderings of different unseen poses for two scenes:  Building Spring (left) and Building Winter (right). We observe fewer errors on ThermoNeRF (highlighted in green) than the baselines. Note that N$_\text{rgb+th}$ stands for Nerfacto$_\text{rgb+th}$.}
  \label{fig:errors-comparison-3}
\end{figure}

Additionally, we visually compare examples of test views rendered by each of the baseline methods as well as by ThermoNeRF (see \cref{fig:results-comparison-four}).
We observe noisy renderings with $\text{Nerfacto}_{\text{th}}$, while Nerfacto$_{\text{rgb+th}}$ generates sensible reconstructions of the geometry.
However, temperature predictions of Nerfacto$_{\text{rgb+th}}$ are influenced by the RGB modality, leading to biased temperature values, sharper edges compared to the ground truth, and elements from RGB images that are visible in the synthesized thermal images.
For example, in \cref{fig:results-comparison-four}, for each Nerfacto$_{rgb+th}$ image, the colors have leaked into the thermal images and buildings are warmer than the ground truth.
While the building (spring) scene (top-row of \cref{fig:results-comparison-four}) looks sharper when synthesized using Nerfacto$_{rgb+th}$ compared to ThermoNeRF, in \cref{fig:errors-comparison} top left row, one can see that the sharpness of the image comes with a high error rate on the temperature rendering---with a MAE of $8.32^\circ$ as opposed to $1.71^\circ$ for ThermoNeRF.
Another example of detail leaking from the RGB into the thermal images can be seen in \cref{fig:errors-comparison} and \cref{fig:errors-comparison-3}, where edges and structures are corresponding to high errors in Nerfacto$_{rgb+th}$.


For in-detail qualitative evaluation of the results, the per-pixel absolute errors in temperature prediction is visualized in \cref{fig:errors-comparison}.
One can see that the highest errors for all models are observed at the edges, due to the ghosting effect of thermal images leading to blurry edges.
It should be noted that ThermoNeRF exhibits the best thermal prediction results while Nerfacto$_{\text{th}}$ shows the worst thermal predictions.
Nerfacto$_{\text{rgb+th}}$ shows higher errors across the scenes than ThermoNeRF due to the contamination of the thermal information by the RGB modality.

Furthermore, cloudy skies are known to provide ideal conditions for infrared measurements outdoors, since the measured object is protected against solar radiation. This fact is supported by the comparison of the thermal image synthesis of Building Spring (top-left) and the Exhibition Building scenes.
During measurements, the Building Spring scene's sky is a mix of clear and cloudy images leading to more noise in the image synthesis than the second scene that consists of a uniform cloudy sky.

\subsection{RGB View Synthesis}

\input{results/rgb_building}

\begin{table}[t]
  \centering
  \setlength{\aboverulesep}{0pt}
  \setlength{\belowrulesep}{0pt}
  \renewcommand{\arraystretch}{1.25}
  \caption{Quantitative comparison of the rendered RGB views between ThermoNeRF and Nerfacto trained only with RGB across the everyday life scenes in ThermoScenes.
  There is no significant degradation by the multimodal learning.
  }

  \resizebox{\textwidth}{!}{%
    \begin{tabular}{llcccccccc>{\columncolor[gray]{0.85}}c}\toprule
      Metric                                                           & Method                  &
      \begin{tabular}[c]{@{}c@{}}Heated \\ Water Cup \end{tabular}     &
      \begin{tabular}[c]{@{}c@{}}Heated \\  Water  Kettle\end{tabular} &
      \begin{tabular}[c]{@{}c@{}}Freezing \\  Ice Cup \end{tabular}    &
      \begin{tabular}[c]{@{}c@{}}Melting \\ Ice  Cup\end{tabular}      &
      \begin{tabular}[c]{@{}c@{}}Double \\  Robot\end{tabular}         &
      \begin{tabular}[c]{@{}c@{}}Raspberry\\  Pi\end{tabular}          &
      Trees                                                            &
      Laptop                                                           &
      Avg                                                                                                                     \\
      \midrule
      \multirow{3}{*}{PSNR $\uparrow$}
                                                                       & $\text{Nerfacto (RGB)}$ &
      18.11                                                            &
      \fs23.00                                                         &
      24.71                                                            &
      \fs17.97                                                         &
      \fs19.05                                                         &
      \fs19.36                                                         &
      20.37                                                            &
      19.3                                                             &
      \fs 20.23
      \\

                                                                       & ThermoNeRF              &
      \fs18.16                                                         &
      21.05                                                            &
      \fs25.69                                                         &
      16.20                                                            &
      19.02                                                            &
      19.07                                                            &
      18.67                                                            &
      18,83                                                            &
      19.58
      \\
                                                                       & Difference              & \textcolor{Green}{+0.05} &
      \textcolor{Red}{-1.95}                                           &
      \textcolor{Green}{+0.98}                                         &
      \textcolor{Red}{-1.77}                                           &
      \textcolor{Red}{-0.03}                                           &
      \textcolor{Red}{-0.29}                                           &
      \textcolor{Red}{-1.61}                                           &
      \textcolor{Red}{-0.47}                                           &
      \textcolor{Red}{-0.63}
      \\
      \midrule
      \multirow{3}{*}{SSIM $\uparrow$}
                                                                       & $\text{Nerfacto (RGB)}$ &
      0.53                                                             &
      \fs0.75                                                          &
      0.73                                                             &
      \fs0.52                                                          &
      \fs0.65                                                          &
      \fs0.73                                                          &
      \fs0.67                                                          &
      \fs 0.61                                                         &
      0.65
      \\

                                                                       & ThermoNeRF              &
      \fs0.56                                                          &
      0.60                                                             &
      \fs0.77                                                          &
      0.43                                                             &
      \fs0.65                                                          &
      0.72                                                             &
      0.59                                                             &
      0.61                                                             &
      0.62                                                                                                                    \\
                                                                       & Difference              & \textcolor{Green}{+0.03} &
      \textcolor{Red}{-0.15}                                           &
      \textcolor{Green}{+0.04}                                         &
      \textcolor{Red}{-0.09}                                           &
      \textcolor{Green}{0.00}                                          &
      \textcolor{Red}{-0.01}                                           &
      \textcolor{Red}{-0.08}                                           &
      \textcolor{Red}{0}                                               &
      \textcolor{Red}{-0.03}

      \\
      \midrule
      \multirow{3}{*}{LPIPS $\downarrow$}
                                                                       & $\text{Nerfacto (RGB)}$ &
      \fs0.25                                                          &
      \fs0.23                                                          &
      0.42                                                             &
      \fs0.17                                                          &
      \fs0.28                                                          &
      \fs0.18                                                          &
      \fs0.32                                                          &
      \fs0.21                                                          &
      0.26

      \\
                                                                       & ThermoNeRF              &
      \fs0.25                                                          &
      0.24                                                             &
      \fs0.37                                                          &
      0.18                                                             &
      0.30                                                             &
      0.19                                                             &
      0.35                                                             &
      \fs0.22                                                          &
      0.26                                                                                                                    \\
                                                                       & Difference              & \textcolor{Green}{0.00}  &
      \textcolor{Red}{+0.01}                                           &
      \textcolor{Green}{-0.05}                                         &
      \textcolor{Red}{+0.01}                                           &
      \textcolor{Red}{+0.02}                                           &
      \textcolor{Red}{+0.01}                                           &
      \textcolor{Red}{+0.03}                                           &
      \textcolor{Red}{+0.01}                                          &
      \textcolor{Red}{0.005}
      \\

      \bottomrule
    \end{tabular}}

  \label{tab:rgb-results}
\end{table}

We investigate whether there is a significant degradation in the quality of RGB views compared with the standard Nerfacto trained only on the RGB inputs.
\cref{tab:rgb-results-buildings} and \cref{tab:rgb-results} summarize the results, showing little to no degradation in the quality of the RGB views.
Indeed, the differences between the PSNR, SSIM and LPIPS of Nerfacto and ThermoNeRF in RGB are respectively $-0.09$, $-0.01$ and $-0.01$ on building scenes, and $-0.63$, $-0.03$ and $0.005$ on other scenes.
The results support that using ThermoNeRF leads to higher accuracy in temperature without a significant degradation in the quality of the reconstructed RGB views. Thus, our approach effectively learns to model 3D scenes across both RGB and thermal modalities, making it suitable for various downstream tasks including building retrofit, energy consumption analysis and non-destructive infrastructure inspection.

%% file: results/outdoor.tex
\begin{table}[t]
  \centering
  \setlength{\aboverulesep}{0pt}
  \setlength{\belowrulesep}{0pt}
  \renewcommand{\arraystretch}{1.25}
  \caption{Quantitative comparison of our method (ThermoNeRF) versus $\text{Nerfacto}_{\text{th}}$ and $\text{Nerfacto}_{\text{rgb+th}}$ on thermal novel view synthesis across the building scenes in ThermoScenes.}

  \resizebox{\textwidth}{!}{%
    \begin{tabular}{llcccccccc>{\columncolor[gray]{0.85}}c}\toprule
      Metric                                                         & Method                            &
      \begin{tabular}[c]{@{}c@{}}Building \\  (Spring) \end{tabular} &
      \begin{tabular}[c]{@{}c@{}}Building \\  (Winter)\end{tabular}  &
      \begin{tabular}[c]{@{}c@{}}Exhibition \\ Building \end{tabular}
                                                                     &
      MED
                                                                     &
      \makecell{Dorm 1}
                                                                     &
      \makecell{Dorm 2}
                                                                     &
      INR
                                                                     &
      BI
                                                                     &
      Avg                                                                                                  \\

      \midrule
      \multirow{3}{*}{\makecell{PSNR                                                                       \\ $\uparrow$}}

                                                                     & $\text{Nerfacto}_{\text{th}}$     &
      20.30                                                          &
      22.80                                                          &
      23.88                                                          &
      19.26                                                          &
      18.65                                                          &
      19.42                                                          &
      18.8                                                           &
      22.95                                                          &
      20.77
      \\

                                                                     & $\text{Nerfacto}_{\text{rgb+th}}$ &
      20.60                                                          &
      28.47                                                          &
      27.74                                                          &
      20.58                                                          &
      32.44                                                          &
      26.64                                                          &
      21.05                                                          &
      27.15                                                          &
      26.24
      \\

                                                                     & ThermoNeRF                        &
      \fs26.63                                                       &
      \fs28.75                                                       &
      \fs33.79                                                       &
      \fs22.04                                                       &
      \fs34.10                                                       &
      \fs29.94                                                       &
      \fs22.45                                                       &
      \fs28.83                                                       &
      \fs 28.62
      \\
      \midrule
      \multirow{3}{*}{\makecell{SSIM                                                                       \\ $\uparrow$}}

                                                                     & $\text{Nerfacto}_{\text{th}}$     &
      0.91                                                           &
      0.87                                                           &
      0.94                                                           &
      0.93                                                           &
      0.92                                                           &
      0.92                                                           &
      0.87                                                           &
      0.85                                                           &
      0.90
      \\

                                                                     & $\text{Nerfacto}_{\text{rgb+th}}$ &
      0.89                                                           &
      \fs0.89                                                        &
      0.95                                                           &
      0.93                                                           &
      0.95                                                           &
      \fs0.95                                                        &
      0.86                                                           &
      0.86                                                           &
      0.91
      \\

                                                                     & ThermoNeRF                        &
      \fs0.92                                                        &
      0.88                                                           &
      \fs0.97                                                        &
      0.93                                                           &
      \fs0.96                                                        &
      \fs0.95                                                        &
      \fs0.88                                                        &
      \fs0.87                                                        &
      \fs 0.92
      \\
      \midrule
      \multirow{3}{*}{\makecell{MAE$_{\text{roi}}$                                                         \\ $^\circ$C $\, \downarrow$}}
                                                                     & $\text{Nerfacto}_{\text{th}}$     &
      6.36                                                           &
      1.80                                                           &
      1.29                                                           &
      4.95                                                           &
      4.37                                                           &
      3.93                                                           &
      1.81                                                           &
      1.70                                                           &
      3.09
      \\
                                                                     & $\text{Nerfacto}_{\text{rgb+th}}$ &
      6.54                                                           &
      0.86                                                           &
      1.00                                                           &
      3.62                                                           &
      0.63                                                           &
      0.77                                                           &
      1.31                                                           &
      1.04                                                           &
      1.79
      \\
                                                                     & ThermoNeRF                        &
      \fs1.88                                                        &
      \fs0.66                                                        &
      \fs0.31                                                        &
      \fs3.10                                                        &
      \fs0.38                                                        &
      \fs0.75                                                        &
      \fs1.28                                                        &
      \fs0.74                                                        &
      \fs 1.04
      \\
      \midrule
      \multirow{3}{*}{\makecell{MAE                                                                        \\ $^\circ$C $\, \downarrow$}}
                                                                     & $\text{Nerfacto}_{\text{th}}$     &
      6.74                                                           &
      1.87                                                           &
      1.56                                                           &
      4.62                                                           &
      4.06                                                           &
      3.65                                                           &
      2.16                                                           &
      1.81                                                           &
      3.11
      \\
                                                                     & $\text{Nerfacto}_{\text{rgb+th}}$ &
      6.59                                                           &
      0.89                                                           &
      0.97                                                           &
      3.80                                                           &
      0.68                                                           &
      3.65                                                           &
      1.63                                                           &
      1.00                                                           &
      2.17
      \\

                                                                     & ThermoNeRF                        &
      \fs 2.40                                                       &
      \fs0.76                                                        &
      \fs0.35                                                        &
      \fs3.38                                                        &
      \fs0.46                                                        &
      \fs0.77                                                        &
      \fs0.93                                                        &
      \fs0.76                                                        &
      \fs 1.13
      \\
      \bottomrule
    \end{tabular}}
  \label{tab:thermal-results-outdoor}
\end{table}

%% file: results/indoor.tex
\begin{table}[t]
  \centering
  \setlength{\aboverulesep}{0pt}
  \setlength{\belowrulesep}{0pt}
  \renewcommand{\arraystretch}{1.25}
  \caption{Quantitative comparison of our method (ThermoNeRF) versus $\text{Nerfacto}_{\text{th}}$ and $\text{Nerfacto}_{\text{rgb+th}}$ on thermal novel view synthesis across the non-building scenes in ThermoScenes.}

  \resizebox{\textwidth}{!}{%
    \begin{tabular}{llcccccccc>{\columncolor[gray]{0.85}}c}\toprule
      Metric                                                           & Method                            &
      \begin{tabular}[c]{@{}c@{}}Heated \\ Water Cup \end{tabular}     &
      \begin{tabular}[c]{@{}c@{}}Heated \\  Water  Kettle\end{tabular} &
      \begin{tabular}[c]{@{}c@{}}Freezing \\  Ice Cup \end{tabular}    &
      \begin{tabular}[c]{@{}c@{}}Melting \\ Ice  Cup\end{tabular}      &
      \begin{tabular}[c]{@{}c@{}}Double \\  Robot\end{tabular}         &
      \begin{tabular}[c]{@{}c@{}}RPi\end{tabular}                      &
      Laptop                                                           &
      Trees                                                            &
      Avg                                                                                                    \\

      \midrule
      \multirow{3}{*}{\makecell{PSNR                                                                       \\ $\uparrow$}}

                                                                       & $\text{Nerfacto}_{\text{th}}$     &
      23.68                                                            &
      29.25                                                            &
      23.34                                                            &
      18.50                                                            &
      10.49                                                            &
      18.08                                                            &
      18.28                                                            &
      20.91                                                            &
      20.32
      \\

                                                                       & $\text{Nerfacto}_{\text{rgb+th}}$ &
      29.76                                                            &
      31.80                                                            &
      22.9                                                             &
      \fs32.70                                                         &
      29.82                                                            &
      24.30                                                            &
      24.28                                                            &

      \fs31.48                                                         &
      28.38
      \\

                                                                       & ThermoNeRF                        &
      \fs32.05                                                         &
      \fs34.04                                                         &
      \fs30.67                                                         &
      32.24                                                            &
      \fs30.75                                                         &
      \fs31.80                                                         &
      \fs25.40                                                         &

      31.07                                                            &
      \fs31.00
      \\
      \midrule
      \multirow{3}{*}{\makecell{SSIM                                                                       \\ $\uparrow$}}

                                                                       & $\text{Nerfacto}_{\text{th}}$     &
      0.71                                                             &
      0.89                                                             &
      0.95                                                             &
      0.93                                                             &
      0.45                                                             &
      0.71                                                             &
      0.65                                                             &

      0.92                                                             &
      0.78
      \\

                                                                       & $\text{Nerfacto}_{\text{rgb+th}}$ &
      0.83                                                             &
      0.91                                                             &
      0.96                                                             &
      \fs0.98                                                          &
      0.89                                                             &
      0.82                                                             &
      0.73                                                             &

      \fs0.94                                                          &
      0.88
      \\

                                                                       & ThermoNeRF                        &
      \fs0.92                                                          &
      \fs0.94                                                          &
      \fs0.97                                                          &
      \fs0.98                                                          &
      \fs0.95                                                          &
      \fs0.96                                                          &
      \fs 0.74                                                         &

      \fs0.94                                                          &
      \fs 0.92
      \\
      \midrule
      \multirow{3}{*}{\makecell{MAE$_{\text{roi}}$                                                         \\ $^\circ$C $\, \downarrow$}}

                                                                       & $\text{Nerfacto}_{\text{th}}$     &
      13.57                                                            &
      5.18                                                             &
      6.75                                                             &
      12.27                                                            &
      2.85                                                             &
      4.82                                                             &
      11.33                                                            &

      1.60                                                             &
      7.30
      \\
                                                                       & $\text{Nerfacto}_{\text{rgb+th}}$ &
      5.35                                                             &
      3.25                                                             &
      10.33                                                            &
      \fs1.26                                                          &
      1.06                                                             &
      1.62                                                             &
      0.54                                                             &

      0.31                                                             &
      2.96
      \\
                                                                       & ThermoNeRF                        &
      \fs2.10                                                          &
      \fs2.76                                                          &
      \fs3.26                                                          &
      1.57                                                             &
      \fs0.91                                                          &
      \fs1.28                                                          &
      \fs0.43                                                          &

      \fs0.25                                                          &
      \fs 1.57
      \\
      \midrule
      \multirow{3}{*}{\makecell{MAE                                                         \\ $^\circ$C $\, \downarrow$}}
                                                                       & $\text{Nerfacto}_{\text{th}}$     &
      1.82                                                             &
      4.19                                                             &
      1.67                                                             &
      1.97                                                             &
      5.33                                                             &
      2.25                                                             &
      0.54                                                             &

      1.49                                                             &
      2.41
      \\
                                                                       & $\text{Nerfacto}_{\text{rgb+th}}$ &
      0.87                                                             &
      1.54                                                             &
      2.01                                                             &
      0.46                                                             &
      0.55                                                             &
      1.15                                                             &
      0.33                                                             &

      0.34                                                             &
      0.91
      \\

                                                                       & ThermoNeRF                        &
      \fs 0.53                                                         &
      \fs 0.71                                                         &
      \fs 0.57                                                         &
      \fs 0.29                                                         &
      \fs0.34                                                          &
      \fs0.27                                                          &
      \fs0.28                                                          &

      \fs0.33                                                          &
      \fs 0.41
      \\
      \bottomrule
    \end{tabular}}
  \label{tab:thermal-results-indoor}
\end{table}

%% file: results/rgb_building.tex
\begin{table}[t]
  \centering
  \setlength{\aboverulesep}{0pt}
  \setlength{\belowrulesep}{0pt}
  \renewcommand{\arraystretch}{1.25}
  \caption{Quantitative comparison of the rendered RGB views between ThermoNeRF and a Nerfacto trained only with RGB across the buildings in ThermoScenes.
  There is no observed significant degradation of the quality of RGB views due to the multimodal learning.
  }

  \resizebox{\textwidth}{!}{%
    \begin{tabular}{llcccccccc>{\columncolor[gray]{0.85}}c}\toprule
      Metric                                                           & Method                  &
      \begin{tabular}[c]{@{}c@{}}Building \\  (Spring) \end{tabular}   &
      \begin{tabular}[c]{@{}c@{}}Building \\  (Winter)\end{tabular}    &
      \begin{tabular}[c]{@{}c@{}}Exhibition \\ Building \end{tabular}  &
      MED &
      Dorm 1 &
      Dorm 2 &
      INR &
      BI &
      Avg                                                                                                                     \\
      \midrule
      \multirow{3}{*}{PSNR $\uparrow$}
                                                                       & $\text{Nerfacto (RGB)}$ &
      \fs19.88                                                         &
      \fs20.02                                                         &
      22.18                                                            &
      \fs17.53&
      \fs19.13&
      \fs19.11&
      \fs14.17&
      20.38&
      \fs19.06
      \\

                                                                       & ThermoNeRF              &
      19.30                                                            &
      19.74                                                            &
      \fs22.81                                                         &
      16.40&
      18.70&
      19.20&
      13.39&
      \fs20.90&
      18.80
      \\
                                                                       & Difference              & 
      \textcolor{Red}{-0.58}                                           &
      \textcolor{Red}{-0.28}                                           &
      \textcolor{Green}{+0.83}                                         &
      \textcolor{Red}{-0.13}&
      \textcolor{Red}{-0.43}&
      \textcolor{Green}{+0.09}&
      \textcolor{Red}{-0.78}&
      \textcolor{Green}{+0.52}&
      \textcolor{Red}{-0.09}
      \\
      \midrule
      \multirow{3}{*}{SSIM $\uparrow$}
                                                                       & $\text{Nerfacto (RGB)}$ &
      \fs0.64                                                          &
      \fs0.62                                                          &
      0.64                                                             &
      0.43&
      0.57&
      0.61&
      0.39&
      0.65&
      0.57
      \\

                                                                       & ThermoNeRF              &
      0.63                                                             &
      0.61                                                             &
      \fs0.66                                                          &
      0.38&
      0.55&
      0.61&
      0.35&
      0.69&
      0.56                                                                                                                    \\
                                                                       & Difference              & 
      \textcolor{Red}{-0.02}                                           &
      \textcolor{Red}{-0.01}                                           &
      \textcolor{Green}{+0.02}                                         &
      \textcolor{Red}{-0.05}&
      \textcolor{Red}{-0.02}&
      \textcolor{Green}{+0.00}&
      \textcolor{Red}{-0.04}&
      \textcolor{Green}{+0.04}&
      \textcolor{Red}{-0.01}

      \\
      \midrule
      \multirow{3}{*}{LPIPS $\downarrow$}
                                                                       & $\text{Nerfacto (RGB)}$ &
      \fs0.24                                                          &
      \fs0.34                                                          &
      \fs0.24                                                          &
      0.32&
      0.17&
      0.19&
      0.59&
      0.27&
      0.29

      \\
                                                                       & ThermoNeRF              &
      0.27                                                             &
      \fs0.34                                                          &
      \fs0.24                                                          &
      0.38&
      0.19&
      0.22&
      0.61&
      0.28&
      0.32                                                                                                                \\
                                                                       & Difference              & 
      \textcolor{Red}{+0.03}                                           &
      \textcolor{Green}{0.00}                                          &
      \textcolor{Green}{0.00}                                          &
      \textcolor{Red}{-0.06}&
      \textcolor{Red}{-0.02}&
      \textcolor{Red}{-0.03}&
      \textcolor{Red}{-0.02}&
      \textcolor{Red}{-0.01}&
      \textcolor{Red}{-0.01}
      \\

      \bottomrule
    \end{tabular}}

  \label{tab:rgb-results-buildings}
\end{table}